\documentclass[10pt,twocolumn,letterpaper]{article}

\usepackage{iccv}
\usepackage{times}
\usepackage{epsfig}
\usepackage{graphicx}
\usepackage{amsmath}
\usepackage{amssymb}
\usepackage{caption}
\usepackage{multirow}

\usepackage{booktabs}
\usepackage{bm}
\usepackage{algorithm, algpseudocode}

\usepackage{color}
\usepackage{colortbl}
\usepackage{textcomp}
\usepackage[outercaption]{sidecap}
\usepackage{makecell}
\usepackage{stmaryrd}
\usepackage{wrapfig}
\usepackage{caption}
\usepackage{enumitem}
\usepackage{subcaption}
\usepackage{pifont}  
\usepackage{array}
\usepackage{enumitem}
\usepackage{pifont}
\usepackage{sidecap}
\usepackage{xspace}

\newcommand{\bx}{{\bm x}}
\newcommand{\gen}{{\mathcal{G}}}
\newcommand{\dis}{{\mathcal{D}}}
\newcommand{\btheta}{{\bm \theta}}
\newcommand{\bpsi}{{\bm \psi}}
\newcommand{\bn}{{\bm \delta}} 
\newcommand{\cls}[1]{{\small{\texttt{#1}}}}

\newcommand{\PreserveBackslash}[1]{\let\temp=\\#1\let\\=\temp}
\newcolumntype{C}[1]{>{\PreserveBackslash\centering}p{#1}}
\newcolumntype{R}[1]{>{\PreserveBackslash\raggedleft}p{#1}}
\newcolumntype{L}[1]{>{\PreserveBackslash\raggedright}p{#1}}

\newcommand{\cmark}{\ding{51}}%
\newcommand{\xmark}{\ding{55}}%

\definecolor{Gray}{gray}{0.90}
\definecolor{LightCyan}{rgb}{0.82,0.82,1}

\newcolumntype{a}{>{\columncolor{Gray}}c}
\newcolumntype{b}{>{\columncolor{LightCyan}}c}

\definecolor{color3}{rgb}{0.95,0.95,0.95}
\definecolor{color4}{rgb}{0.96,0.96,0.86}
\definecolor{color1}{rgb}{0.90,0.94,0.84}
\definecolor{color2}{rgb}{1,0.92,0.8}

\newcommand{\ours}[0]{{\textsc{TTP}}\xspace}
\makeatletter
\def\adl@drawiv#1#2#3{%
        \hskip.5\tabcolsep
        \xleaders#3{#2.5\@tempdimb #1{1}#2.5\@tempdimb}%
                #2\z@ plus1fil minus1fil\relax
        \hskip.5\tabcolsep}
\newcommand{\cdashlinelr}[1]{%
  \noalign{\vskip\aboverulesep
           \global\let\@dashdrawstore\adl@draw
           \global\let\adl@draw\adl@drawiv}
  \cdashline{#1}
  \noalign{\global\let\adl@draw\@dashdrawstore
           \vskip\belowrulesep}}
\makeatother

\usepackage[pagebackref=true,breaklinks=true,letterpaper=true,colorlinks,bookmarks=false]{hyperref}

\iccvfinalcopy 

\def\iccvPaperID{10534} 

\ificcvfinal\fi

\begin{document}

\title{On Generating Transferable Targeted Perturbations}
\author{
    Muzammal Naseer$^{*}$, ~~Salman Khan$^{\dagger}$, ~~Munawar Hayat$^ \mathsection$, 
    ~~Fahad Shahbaz Khan$^{\dagger}$, ~~Fatih Porikli$^\ddagger$\\
  $^*$Australian National University, Australia, ~~$^ \mathsection$Monash University, Australia, ~~$^\ddagger$Qualcomm, USA\\
  $^\dagger$Mohamed bin Zayed University of Artificial Intelligence, UAE\\
  \texttt{\small muzammal.naseer@anu.edu.au}, 
~  \texttt{\small \{salman.khan,fahad.khan\}@mbzuai.ac.ae}, ~ \texttt{\small munawar.hayat@monash.edu.au}\\
  \texttt{\small fatih.porikli@gmail.com}
  }


\maketitle
\ificcvfinal\thispagestyle{empty}\fi

\begin{abstract}
  While the untargeted black-box transferability of adversarial perturbations has been extensively studied before, changing an unseen model's decisions to a specific `targeted' class remains a challenging feat. In this paper, we propose a new generative approach for highly transferable targeted perturbations (\ours). We note that the existing methods are less suitable for this task due to their reliance on class-boundary information that changes from one model to another, thus reducing transferability. In contrast, our approach matches the  perturbed image `distribution' with that of the target class, leading to high targeted transferability rates. To this end, we propose a new objective function that not only aligns the global distributions of source and target images, but also matches the local neighbourhood structure between the two domains. Based on the proposed objective, we train a generator function that can adaptively synthesize perturbations specific to a given input. Our generative approach is independent of the source or target domain labels, while consistently performs well against state-of-the-art methods on a wide range of attack settings. As an example, we achieve $32.63\%$ target transferability from (an adversarially weak) VGG19$_{BN}$ to (a strong) WideResNet on ImageNet val. set, which is 4$\times$ higher than the previous best generative attack and 16$\times$ better than instance-specific iterative attack. Code is available at:  {\small\url{https://github.com/Muzammal-Naseer/TTP}}.
\end{abstract}

\vspace{-0.9em}
\section{Introduction}

\begin{figure}[t]
\centering
  	\centering
    \includegraphics[width=\linewidth, keepaspectratio]{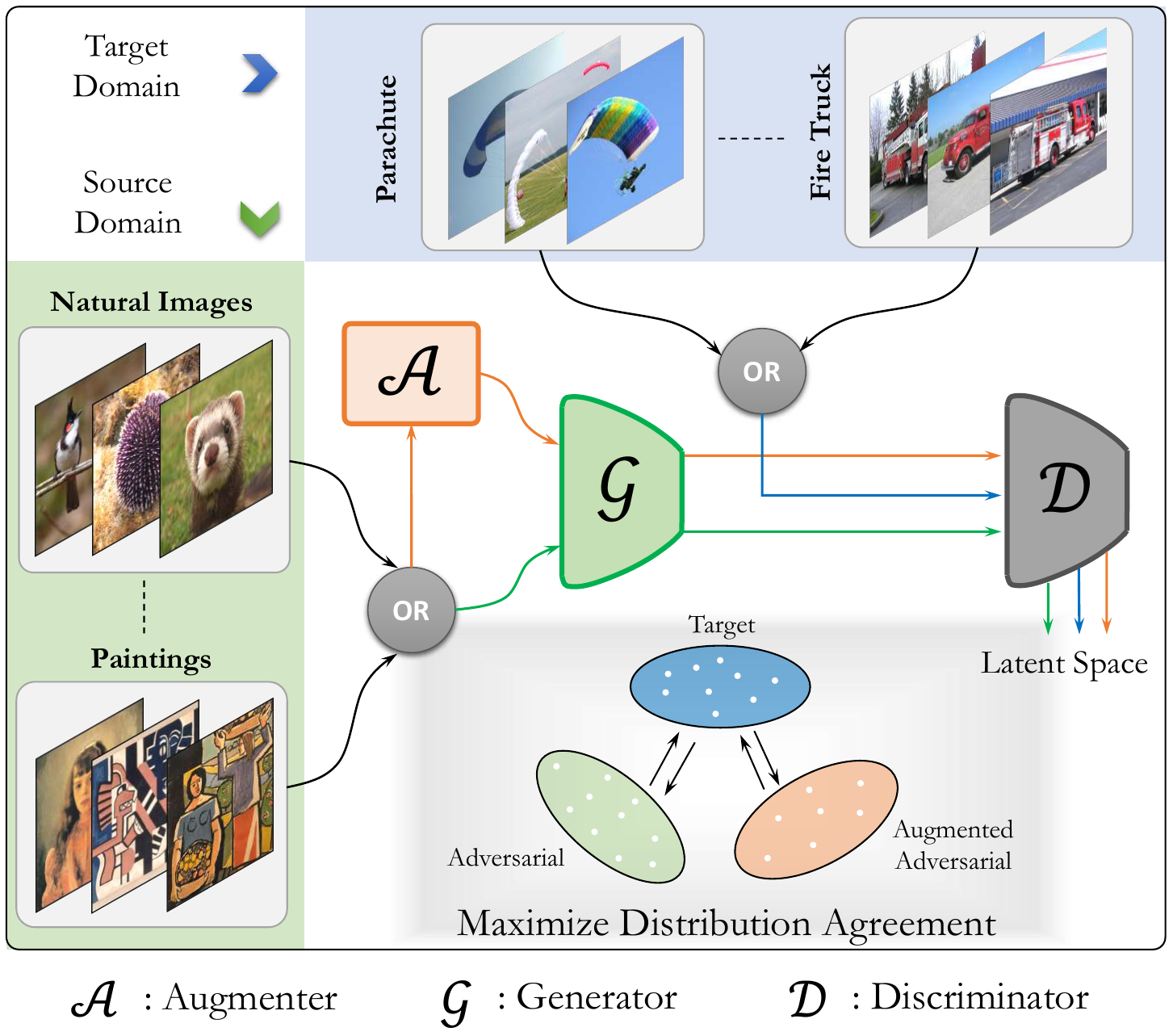}\
    \caption{ Attack Overview (\ours): Instead of finding perturbations specific to a class-boundary information learned by a model, \ours seeks to match global distribution statistics between the source and the target domains. Specifically, our generator function is trained to maximize agreement between the \emph{perturbed} source distribution, its augmented version and the target distribution in the feature space. Importantly, our attack can function in an unsupervised fashion and does not require source domain to be the same as target (e.g., perturbations can be learned from paintings to transfer on natural images). }
\label{fig: demo_fig}
\end{figure}

We study the challenging problem of \emph{targeted} transferability of adversarial perturbations. In this case, given an input sample from any source category, the goal of the adversary is to change the decision of an \emph{unknown} model to a \emph{specific} target class (e.g., misclassify any painting image to \cls{Fire truck}, see Fig.~\ref{fig: demo_fig}). This task is significantly more difficult than merely changing the decision to a random target class or any similar class (\eg, changing `\cls{cat}' to `\cls{aeroplane}' is more difficult than altering the decision to `\cls{dog}'). Target transferability can therefore lead to \emph{goal-driven} adversarial perturbations that provide desired control over the attacked model. However, target transferability remains challenging for the current adversarial attacks \cite{madry2018towards,liu2016delving,dong2018boosting,xie2019improving,inkawhich2019feature,inkawhich2020perturbing,Inkawhich2020Transferable,li2020towards,wu2020skip} that transfer adversarial noise in a \emph{black-box} setting, where architecture and training mechanism of the attacked model remain unknown, and the attack is restricted within a certain perturbation budget.
 
We observe that modest performance of existing methods on targeted transferability is due to their reliance on class-boundary information learned by the model which lacks generalizability. For example, iterative \emph{instance-specific} attacks rely on the classification score information to perturb a given sample, thereby ignoring the global \emph{class-specific} information \cite{madry2018towards,dong2018boosting,xie2019improving,inkawhich2020perturbing}. Such adversarial directions also vary across different models \cite{liu2016delving}, leading to poor target transferability \cite{liu2016delving, dong2018boosting}. On the other hand, although universal and generative perturbations are designed to encode global noise patterns  \cite{moosavi2017universal,poursaeed2018generative,naseer2019cross}, they still exploit the class impressions learned by a neural network which alone are not fully representative of the target distribution, thereby achieving only modest black-box fooling rates \cite{reddy2018ask}. Furthermore, they are dependent on the classification information, necessitating a supervised pretrained model for generator's guidance and therefore cannot directly work with unsupervised features \cite{chen2020simple, he2020momentum}. Another group of techniques exploit intermediate features, but they either find untargeted perturbations by design \cite{mopuri2017fast,reddy2018nag} or are limited in their capacity to transfer targeted perturbations \cite{li2020towards, inkawhich2019feature, Inkawhich2020Transferable,inkawhich2020perturbing}.

We introduce a novel generative training framework which maps a given source distribution to a specific target distribution by maximizing the mutual agreement between the two in the latent space of a pretrained discriminator. 
Our main contributions are:
\begin{itemize}[leftmargin=*,noitemsep]
    \item \emph{\textbf{Generative Targeted Transferability:}} We propose a novel generative approach to learn transferable targeted adversarial perturbations. Our unique training mechanism allows the generator to explore augmented adversarial space during training which enhances the transferability of adversarial examples during inference (Sec.~\ref{subsec:generative_perturbations}).
    
    \item \emph{\textbf{Mutual Distribution Matching:}} Our training approach is based on maximizing the mutual agreement between the given source and the target distribution. Therefore, our method can provide targeted guidance to train the generator without the need of classification boundary information. This allows an attacker to learn targeted generative perturbations from the unsupervised features \cite{chen2020simple, he2020momentum} and eliminate the cost of labelled data (Sec.~\ref{subsec:dist_matching}).
    \item \emph{\textbf{Neighbourhood Similarity Matching:}} Alongside global distribution matching, we introduce batch-wise neighbourhood similarity matching objective between adversarial and target class samples to maximize the local alignment between the two distributions (Sec. \ref{subsec: neighbourhood_similarity_matching}).
\end{itemize}
Our extensive experiments on various ImageNet splits and CNN architectures show state-of-the-art targeted transferability against naturally and adversarially trained models, stylized models and input-processing based defenses. The results demonstrate our benefit compared to recent targeted instance-specific as well as other generative methods. Further, our attack demonstrates rapid convergence.

\section{Related Work}
\noindent\textbf{Iterative Instance-Specific Perturbations:} After Szegedy \etal \cite{szegedy2013intriguing} highlighted the vulnerability of neural networks, many adversarial attacks have been introduced to study if the adversarial examples are transferable from one model to another, when a target model is unknown. Among these, iterative instance-specific attacks \cite{dong2018boosting, xie2019improving, dong2019evading} perturb a given sample by iteratively using gradient information. Target transferability of such attacks is very poor \cite{dong2018boosting,liu2016delving} (as shown in Sec.~\ref{sec: experiments}). Other attacks also use feature space either by maximizing the feature difference \cite{zhou2018transferable, huang2019enhancing, li2020yet} or applying attention \cite{wu2020boosting} or avoiding non-linearity while back-propagating gradients \cite{guo2020backpropagating} or exploiting skip-connections \cite{wu2020skip}. However, these attacks are mainly designed to enhance non-targeted transferability which is an easier problem. Recently, different instance-specific (transferable) targeted attacks have been proposed including \cite{li2020towards} which introduces a triplet loss to push adversarial examples towards the target label while increasing their distance from the original label. Inkawhich \etal \cite{Inkawhich2020Transferable, inkawhich2020perturbing} proposed to exploit feature space \cite{inkawhich2019feature} along with the classifier information \cite{inkawhich2020perturbing} to generate target adversaries that are shown to transfer relatively better than other instance-specific attacks. These attacks \cite{Inkawhich2020Transferable, inkawhich2020perturbing} have the following limitations. \textbf{a)} They need access to a \emph{labeled} dataset \eg, ImageNet~\cite{ILSVRC15} in order to train one-vs-all binary classifiers for attacked target classes. \textbf{b)} They need to identify best performing single layer \cite{Inkawhich2020Transferable} or a combination of layers \cite{inkawhich2020perturbing} which adds further complexity to attack optimization. \textbf{c)} Finally, the attack performance degrades significantly with  quality of features, \eg, it struggles to transfer target perturbations from VGG models \cite{Inkawhich2020Transferable}.

\noindent\textbf{Universal Perturbation:} In contrast to instance-specific perturbations, \cite{moosavi2017universal}  learns a single universal noise pattern which is representative of the entire data distribution and can fool a model on majority of samples. Li \etal \cite{li2019regional} introduce gradient transformation module to find smooth universal patterns while \cite{mopuri2017fast} shows that such patterns can be found without any training data. 
Although universal perturbations \cite{moosavi2017universal, mopuri2018generalizable, mopuri2017fast, li2019regional} based attacks are efficient (the attacker just needs to add the noise to any given sample at inference), they are limited in their capacity to yield transferable adversaries which can generalize across different data distributions and models \cite{poursaeed2018generative, naseer2019cross}.

\noindent\textbf{Generative Perturbations:}  Generative adversarial perturbations perform better than directly optimizing universal noise \cite{reddy2018nag, poursaeed2018generative,naseer2019cross}. Poursaeed \etal \cite{poursaeed2018generative} proposed the first generative approach to adapt perturbations to an input sample. Naseer \etal \cite{naseer2019cross} improved this framework with relativistic training objective which also allows cross-domain transferability.  Our method belongs to the generative category and can adapt to an input sample with a single forward pass.  Unlike \cite{reddy2018nag, poursaeed2018generative, naseer2019cross}, we seek to fool the model by matching \emph{distributions} of source and targets with distribution matching and neighbourhood similarity criteria. Our proposed framework does not require labeled source or target data and can extract target perturbations from a discriminator model trained in an unsupervised manner while previous generative methods are dependent on class-boundary information learned by the model. Further, our method converges faster (Sec.~\ref{sec: experiments}) and provides improved targeted  transferability owing to its novel loss and training mechanism.

\begin{figure*}[t]
\centering
  	\centering
    \includegraphics[ width=0.95\linewidth, keepaspectratio]{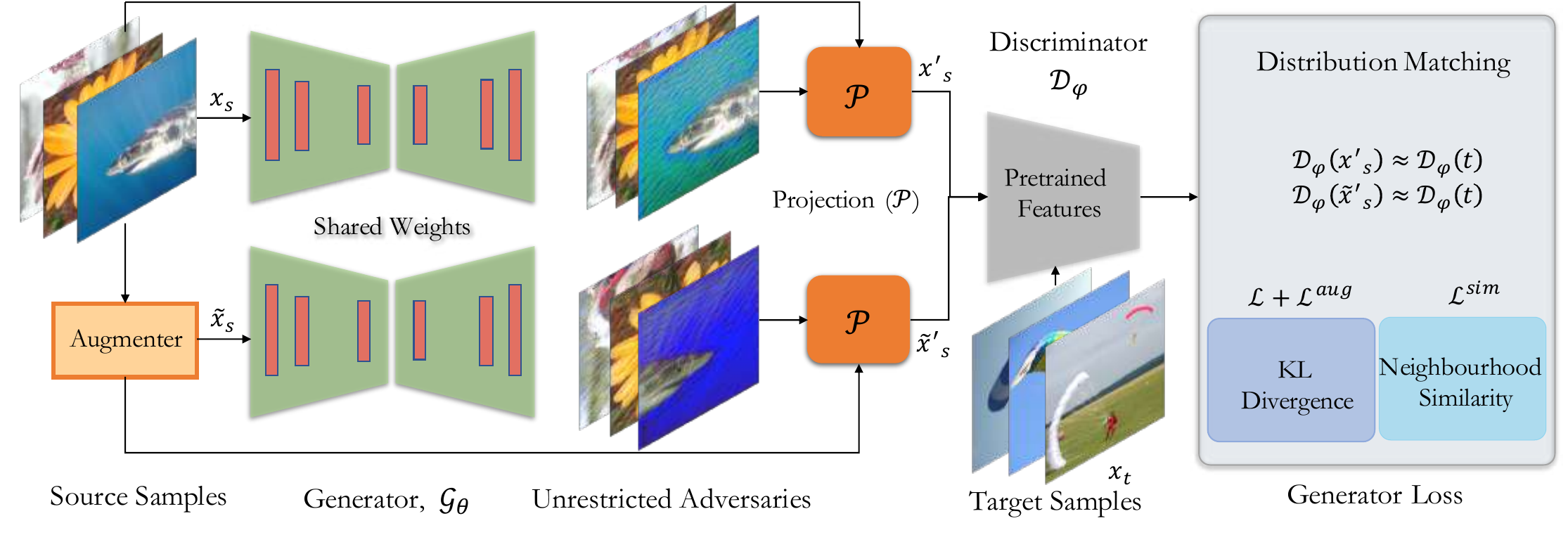}\
    \caption{\emph{Targeted Transferable Perturbations:} During training, \ours matches adversarial and
     augmented adversarial samples to a target domain within discriminator's latent space for improved transferability. The adversarial samples corresponding to original and augmented images are bounded (via projection) around their source samples to explore adversarial space around natural as well as augmented samples.}\vspace{-1em}
\label{fig: concept_fig}
\end{figure*}

\section{Generating Targeted Adversaries}
Our goal is to craft adversarial perturbations $\delta$ that can fool a model to misclassify any given input to a specific target class $t$. We assume access to source and target domain data represented by  $P$ and $Q$, from which the source and target class samples are obtained i.e., $\bx_s \sim P$, $\bx_t \sim Q$. The source and target domains are likely to be non-aligned i.e., $P \neq Q$, making it challenging to achieve targeted transferability of adversarial  perturbations.  We also consider a perturbed source data $P'$ that comprises of adversarially manipulated samples ${\bx}'_s \sim P'$ where ${\bx}'_s = \bx_s + \delta$. $\bx_s$, $\bx'_s$ and $\bx_t$ represent source, adversarial and target domain samples while  $\dis_{\bpsi} (\bx_s)$, $\dis_{\bpsi} (\bx'_s)$ and $\dis_{\bpsi} (\bx_t)$ are their corresponding latent distributions.

\subsection{Generative Model}\label{subsec:generative_perturbations}
We propose a generative approach to perturb the source domain samples $\bx_s$ to a specified target class. The framework (see Fig.~\ref{fig: concept_fig}) consists of a generator $\gen_{\btheta}$ and a discriminator $\dis_{\bpsi}$ parameterized by $\btheta$ and $\bpsi$, respectively. The generator function $\gen_{\btheta}$ learns a mapping from the source images to the target category such that the input images are minimally changed i.e., adversarial noise $\delta$ is strictly constrained under a norm distance $l_\infty \le \epsilon$.
This is ensured by projecting the unbounded adversaries from $\gen_{\btheta}$ within fixed norm distance of $\bx_s$ using a differentiable clipping operation,
\begin{equation}\label{eq: smooth projection}
    {\bx}'_s = \text{clip}\big(\min (\bx_s+\epsilon, \max (\mathcal{W} * \gen_{\btheta}(\bx_s),\bx_s-\epsilon))\big),
\end{equation}
where, $\mathcal{W}$ is a smoothing operator with fixed weights that reduces high frequencies without violating the $l_\infty$ distance constraint.  The smooth projection in Eq.~\ref{eq: smooth projection} (denoted by $\mathcal{P}$ in Fig.~\ref{fig: concept_fig}) not only tightly bounds generator's output within $l_\infty$ norm but also encourages avoiding redundant high frequencies \cite{olah2017feature} during the optimization process. This allows the generator to converge to a more meaningful solution. 

The existing generative designs for adversarial attacks \cite{poursaeed2018generative, naseer2019cross} leverage the decision space of the discriminator to craft perturbations. In such cases, the \emph{class-boundary} information learned by the discriminator is used to fool DNN models (\eg for ImageNet, discriminator is pretrained on 1k classes). This dependence is problematic since an attacker must have access to a discriminator trained on large-scale labeled dataset \cite{deng2009imagenet}. Attacker then tries to learn target class impressions using either cross-entropy (CE) \cite{poursaeed2018generative} or relativistic CE \cite{naseer2019cross}. Thus, the generated perturbations are directly dependent on the quality of the discriminator's classification space. Furthermore, the generated adversaries are dependent on the input instance-specific features and do not model the global properties of the target distribution, resulting in only limited transferability.

To address above limitations, our generative design models the target distribution $Q$ and pushes the perturbed source distribution $P'$ closer to $Q$ using the latent space of $\dis_{\bpsi}$, 
\begin{align}\label{eq: main_objective}
     \parallel \bn \parallel_{\infty} \leq \epsilon, \quad s.t., \quad   \dis_{\bpsi} ({\bx}'_s) \approx \dis_{\bpsi} (\bx_t).
\end{align}
This global objective provides two crucial benefits. \emph{First,} reducing mismatch between perturbed and target distributions provides an improved guidance to the generator. The resulting perturbations well align the input samples with the target distribution, leading to transferable adversaries. \emph{Second,} the distributions alignment task makes us independent of the $\dis_{\bpsi}$'s classification information. In turn, our approach can function equally well with a discriminator trained in a self-supervised manner on unlabelled data \cite{chen2020simple, he2020momentum}. In our case, we simply align the feature distributions from $\dis_{\bpsi}$ to match $P'$ and $Q$. Thus, for a given sample $\bx$, n-dimensional features are obtained  i.e., $\dis_{\bpsi}(\bx) \in \mathbb{R}^n$. If $\dis_{\bpsi}$ is trained in a supervised manner on ImageNet then $n=1000$, and if $\dis_{\bpsi}$ is trained in an unsupervised fashion then $n$ is equal to the output feature dimension. 

\begin{algorithm}[t]
\small
\caption{Generating \ours}
\label{alg: TTP}
\begin{algorithmic}[1]

\Require Source data $\mathcal{X}_s$, Target data $\mathcal{X}_t$,  pretrained discriminator $\dis_{\bpsi}$, perturbation budget $\epsilon$, loss criteria $\mathcal{L}_{\mathcal{G}}$.

\Ensure Randomly initialize the generator, $\gen_{\btheta}$
\Repeat
\State Randomly sample mini-batches $\bx_s \sim \mathcal{X}_s$  and $\bx_t \sim \mathcal{X}_t$
\State Create augmented copy of the source mini-batch $\tilde{\bx}_s$.
\State Forward-pass $\bx_s$ and $\tilde{\bx}_s$ through the generator and 
\Statex \hspace{0.4cm} generate unbounded adversaries; $\bx'_s$, $\tilde{\bx}'_s$. 
\State Bound the adversaries using Eq.~\ref{eq: smooth projection} such that:
\begin{align*}
      \| \bx'_s - \bx_s\|_{\infty} \le \epsilon \;\; \text{and} \;\;
       \| \tilde{\bx}'_s - \tilde{\bx}_s\|_{\infty} \le \epsilon
\end{align*}
  
\State Forward pass $\bx'_s$, $\tilde{\bx}'_s$ and $\bx_t$ through $\dis_{\bpsi}$.
\State Compute the matching losses; $\mathcal{L}$, $\mathcal{L}^{aug}$ and $\mathcal{L}^{sim}$ using \Statex \hspace{0.4cm} Eq. \ref{eq: loss_on_source}, \ref{eq: loss_on_aug} and \ref{eq: loss_sim}, respectively.
\State Compute the generator loss given in Eq. \ref{eq: overall_loss}.  
\State Backward pass and update $\gen_{\btheta}$
\Until{ $\gen_{\btheta}$ converges.}
\end{algorithmic}
\end{algorithm}

\vspace{0.8em}
\subsection{Distribution Matching}\label{subsec:dist_matching}
We measure the mutual agreement between $P'$ and $Q$ using Kullback Leibler (KL) divergence defined on discriminator features $\dis_{\bpsi}({\bx}'_s)$ and $ \dis_{\bpsi}({\bx}_t)$,
\begin{align}
\resizebox{\linewidth}{!}{$ 
    D_{KL}(P'\|Q)  = \frac{1}{N} \sum\limits_{i=1}^N \sum\limits_{j=1}^n \sigma(\dis_{\bpsi}(\bx'^{,i}_{s}))_j \log\frac{\sigma(\dis_{\bpsi}(\bx'^{,i}_{s}))_{j}}{\sigma(\dis_{\bpsi}(\bx^{i}_{t}))_{j}}, \notag
    $}
\end{align}
where $N$ represents the number of samples, $n$ is the discriminator's output dimension, and $\sigma$ denotes the softmax operation. In simple terms, KL divergence measures the difference between two distributions in terms of the average surprise in experiencing $\bx_t$ when we expected to see $\bx'_s$. Since KL divergence is asymmetric \ie $D_{KL}(P'\|Q) \neq D_{KL}(Q \| P')$, and not a valid distance measure, we define our loss function for distribution matching \cite{kullback1951information} as follows:
\begin{equation}\label{eq: loss_on_source}
    \mathcal{L} = D_{KL}(P'\|Q) + D_{KL}(Q\|P').
\end{equation}

As a regularization measure, we add augmented versions of the source domain samples during distribution matching. This enables the generator to focus specifically on adding target class-specific patterns that are robust to input transformations. To this end, we randomly apply rotation, crop resize, horizontal flip, color jittering or gray scale transformation to create augmented samples $\tilde{\bx}_s$ from the original $\bx_s$. The $\tilde{\bx}_s \sim \tilde{P}$ are passed through the $\gen_{\theta}$ and the perturbed augmented samples $\tilde{\bx}'_s \sim \tilde{P}'$ are projected using Eq.~\ref{eq: smooth projection} to stay close to the augmented samples i.e.,  $\| \tilde{\bx}'_s - \tilde{\bx}_s\|_{\infty} \le \epsilon$. No augmentation is applied to the target domain samples. We then pass $\tilde{\bx}'_s$ through the discriminator and compute the mutual agreement between $\dis_{\bpsi} (\tilde{\bx}'_s)$ and 
$ \dis_{\bpsi}({\bx}_t)$ as follows:
\begin{align}\label{eq: loss_on_aug}
     \mathcal{L}^{aug} &= D_{KL}(\tilde{P}' \| Q) + D_{KL}(Q \| \tilde{P}').
\end{align}
The impact of data augmentations and their effectiveness for our proposed targeted attack is studied in Sec.~\ref{sec: experiments}.

\subsection{Neighbourhood Similarity Matching}\label{subsec: neighbourhood_similarity_matching}
The above objective promotes alignment between the distributions but does not consider the local structure e.g., the relationship between a sample and its augmented versions. For a faithful alignment between perturbed source samples and the target class samples, we propose to also match the \emph{neighbourhood similarity distributions} between the two domains. Specifically, consider a batch of target domain samples $\{\bx^i_t\}_{i=1}^{N}$ and a batch of perturbed source domain samples $\{\bx'^i_s\}_{i=1}^{N}$. For the case of $\bx'_s$,  in a given training batch, we compute a similarity matrix $\mathcal{S}^s$ whose elements encode the cosine similarity between the original sample and its augmented version $\tilde{\bx}'_s$, i.e.,
\begin{align}
\small
    \mathcal{S}^s_{i,j} = \frac{  \dis_{\bpsi}(\bx'^{,i}_s) \boldsymbol{\cdot} \dis_{\bpsi}(\tilde{\bx}'^{,j}_s) }{ \| \dis_{\bpsi}(\bx'^{,i}_s) \| \| \dis_{\bpsi}(\tilde{\bx}'^{, j}_s) \| }.
\end{align}
In contrast, for the case of $\bx_t$, we compute similarity between only the original target samples (no augmentations) as we need to model the local neighbourhood connectivity in the target domain. This choice is impractical for the source domain case where many categories co-exist, while for the target distribution, we assume a single category.  Thus the target similarity matrix $\mathcal{S}^t$ is computed as,
\begin{align}
    \mathcal{S}^t_{i,j} = \frac{  \dis_{\bpsi}(\bx^i_t) \boldsymbol{\cdot} \dis_{\bpsi}(\bx_t^{j}) }{ \| \dis_{\bpsi}(\bx^i_t) \| \| \dis_{\bpsi}(\bx_t^{j}) \| }.
\end{align}
The resulting similarity matrices are normalized along the row dimension with softmax to obtain probability estimates, 
\begin{align}
    \bar{\mathcal{S}}_{i,j} = \frac{\exp(\mathcal{S}_{i,j})}{ \sum_k  \exp(\mathcal{S}_{i,k})}, \text{ where, } \mathcal{S} \in \{\mathcal{S}^s, \mathcal{S}^t \}.
\end{align}
Here, each term shows the probability with which the two sample pairs are related to each other. Given $\bar{\mathcal{S}}^s$ and $\bar{\mathcal{S}}^t$, we compute the KL divergence to enforce a loss term that seeks to match the local neighbourhood patterns between source and target domains,
\begin{align}\label{eq: loss_sim}
    \mathcal{L}^{sim} =  \sum_{i,j} \bar{\mathcal{S}}^t_{i,j} \log \frac{\bar{\mathcal{S}}^t_{i,j}}{\bar{\mathcal{S}}^s_{i,j}} + \sum_{i,j} \bar{\mathcal{S}}^s_{i,j} \log \frac{\bar{\mathcal{S}}^s_{i,j}}{\bar{\mathcal{S}}^t_{i,j}}.
\end{align}

\subsection{Overall loss function}
Finally, the generator parameters are updated by minimizing the following loss (Algorithm \ref{alg: TTP}):
\begin{equation}\label{eq: overall_loss}
    \mathcal{L}_{\mathcal{G}} = \mathcal{L} + \mathcal{L}^{aug} + \mathcal{L}^{sim}.
\end{equation}
This loss encourages the generator to perturb source samples that not only match the global characteristics of the target distribution ($\mathcal{L}+ \mathcal{L}^{aug}$), but also the local information based on neighbourhood connectivity ($\mathcal{L}^{sim}$).

\section{Experiments}\label{sec: experiments}

\begin{figure*}[t]
\small \centering
\begin{minipage}{.7\textwidth}
\vspace{-2em}
    \begin{minipage}{.16\textwidth}
  	\centering
  	  Indigo Bunting
    \includegraphics[ width=\linewidth, height=1.6cm]{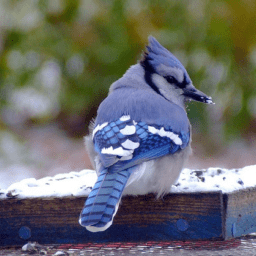}\
  \end{minipage}
    \begin{minipage}{.16\textwidth}
  	\centering
  	  Cardoon
    \includegraphics[ trim= 0mm 0mm 0mm -2mm, width=\linewidth, height=1.6cm]{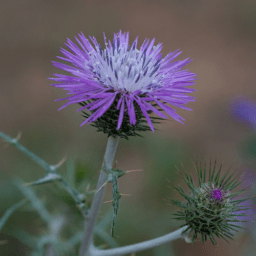}\
  \end{minipage}
    \begin{minipage}{.16\textwidth}
  	\centering
  	  Impala
    \includegraphics[ width=\linewidth, height=1.6cm]{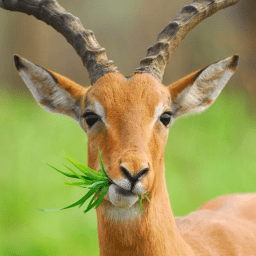}\
  \end{minipage}
    \begin{minipage}{.16\textwidth}
  	\centering
  	Wood Rabbit
    \includegraphics[trim= 0mm 0mm 0mm -2mm, width=\linewidth, height=1.6cm]{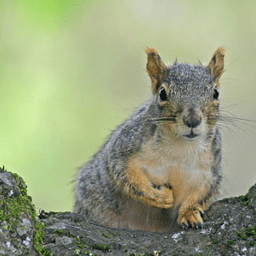}\
  \end{minipage}
\begin{minipage}{.16\textwidth}
  	\centering
  	  Crane
    \includegraphics[trim= 0mm 0mm 0mm -2mm,width=\linewidth, height=1.6cm]{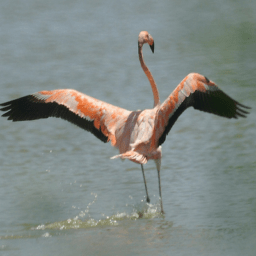}\
  \end{minipage}
  \begin{minipage}{.16\textwidth}
  	\centering
  	  Elephant
    \includegraphics[width=\linewidth, height=1.6cm]{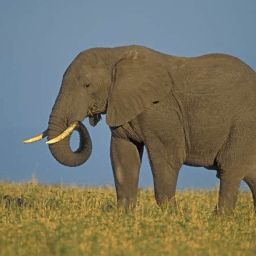}\
  \end{minipage}
  \\
  \begin{minipage}{.16\textwidth}
  	\centering
    \includegraphics[ width=\linewidth, height=1.6cm]{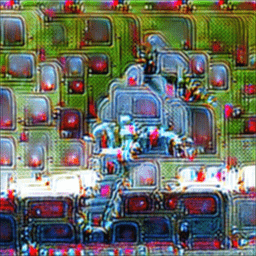}\
  \end{minipage}
    \begin{minipage}{.16\textwidth}
  	\centering
    \includegraphics[width=\linewidth, height=1.6cm]{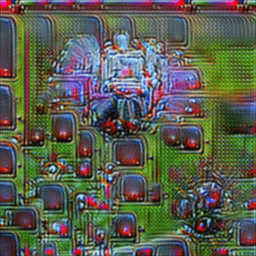}\
  \end{minipage}
    \begin{minipage}{.16\textwidth}
  	\centering
    \includegraphics[width=\linewidth, height=1.6cm]{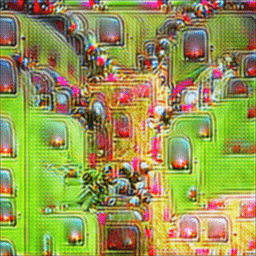}\
  \end{minipage}
    \begin{minipage}{.16\textwidth}
  	\centering
    \includegraphics[width=\linewidth, height=1.6cm]{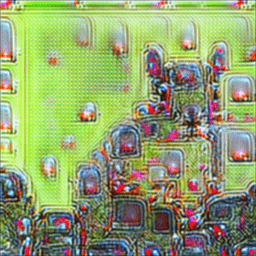}\
  \end{minipage}
\begin{minipage}{.16\textwidth}
  	\centering
    \includegraphics[width=\linewidth, height=1.6cm]{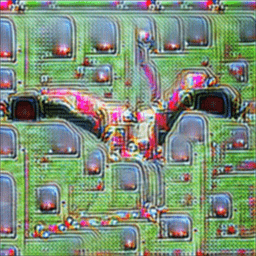}\
  \end{minipage}
  \begin{minipage}{.16\textwidth}
  	\centering
    \includegraphics[width=\linewidth, height=1.6cm]{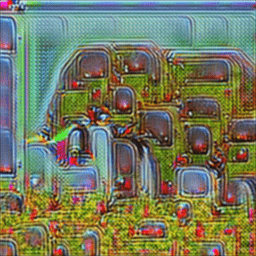}\
  \end{minipage}
  \\
  \begin{minipage}{.16\textwidth}
  	\centering
    \includegraphics[ width=\linewidth, height=1.6cm]{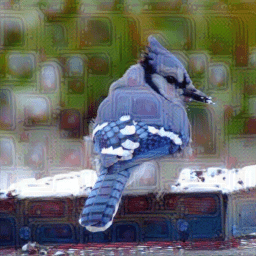}\
     Fire Truck
  \end{minipage}
    \begin{minipage}{.16\textwidth}
  	\centering
    \includegraphics[width=\linewidth, height=1.6cm]{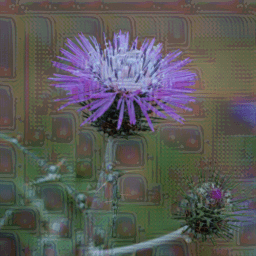}\
     Fire Truck
  \end{minipage}
    \begin{minipage}{.16\textwidth}
  	\centering
    \includegraphics[width=\linewidth, height=1.6cm]{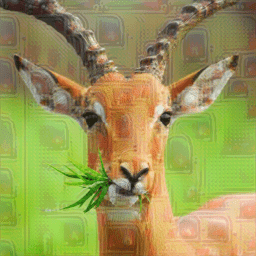}\
     Fire Truck
  \end{minipage}
    \begin{minipage}{.16\textwidth}
  	\centering
    \includegraphics[width=\linewidth, height=1.6cm]{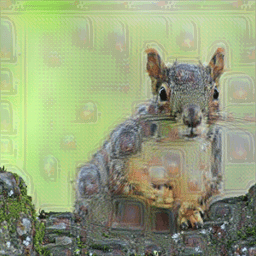}\
     Fire Truck
  \end{minipage}
\begin{minipage}{.16\textwidth}
  	\centering
    \includegraphics[width=\linewidth, height=1.6cm]{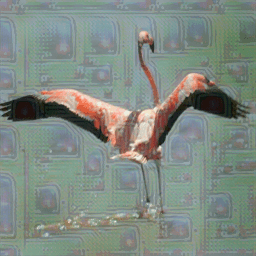}\
     Fire Truck
  \end{minipage}
  \begin{minipage}{.16\textwidth}
  	\centering
    \includegraphics[width=\linewidth, height=1.6cm]{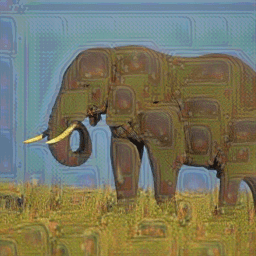}\
     Fire Truck
  \end{minipage}
     
\end{minipage}
 \hfill
\begin{minipage}{.28\textwidth}
\captionof{figure}{Targeted adversaries produced by a TTP generator learned to maximize the agreement with  'Fire Truck' distribution against Dense121 ImageNet model. 1st and 2nd rows show clean images and unrestricted outputs of the adversarial generator, respectively. 3rd row shows adversaries after valid projection. See \ref{sec: sup_visual_demo} for more qualitative examples including comparisons between targeted patterns learned by TTP from different source models of a certain family of networks.}
\label{fig:images}
\end{minipage}
\end{figure*}

Our generator $\gen_{\btheta}$ is based on ResNet architecture \cite{Johnson2016PerceptualLF}, and outputs an adversarial sample with the same size as of input (Fig. \ref{fig:images}). This generator architecture is the same as in the baseline generative attacks \cite{poursaeed2018generative, naseer2019cross}. Our discriminator $\dis_{\bpsi}$ is pre-trained in a supervised or self-supervised manner. For training $\gen_{\btheta}$, we freeze $\dis_{\bpsi}$. We use Adam optimizer \cite{kingma2014adam} with a learning rate of $10^{-4}$ ($\beta_1=.5, \beta_2=.999$) for 20 epochs. For source domain data, we use 50k random images from ImageNet train set. Our method is not sensitive to the choice of source samples since it can learn transferable perturbations even from other domains \eg Paintings. Similar to other generative methods \cite{poursaeed2018generative, naseer2019cross}, we fix source data. For target domain data, we use $1300$ images for each target collected from ImageNet training set (without their original labels). We used default settings or implementations as provided by the authors of baseline attacks. Similarly, we used open-sourced (pretrained) stylized \cite{geirhos2018imagenettrained}, adversarial \cite{salman2020adversarially} and purifier (NRP) \cite{Naseer_2020_CVPR} models to evaluate robustness.

\subsection{Evaluation Settings}
We perform inference on ImageNet validation set (50k samples). \emph{No augmentations are applied at inference time}. The perturbation budget is tightly bounded and clearly mentioned in each experiment following the standard practices $l_\infty \le 16$ \cite{dong2018boosting, inkawhich2019feature, inkawhich2020perturbing} and $l_\infty \le 32$ \cite{naseer2019cross, li2019regional}. We perturb all the ImageNet val. samples (except the target samples) to the pre-defined target class. We repeat this process for all the given targets and report Top-1 (\%) accuracy averaged across all targets. We compare our method under two main settings (10-Targets and 100-Targets), as described below. 

\noindent\textbf{10-Targets:}
 We further consider two settings. \textbf{(a)} \textit{10-Targets (subset-source)} which is consistent with \cite{Inkawhich2020Transferable} and has a subset of source classes at inference. \textbf{(b)} \textit{10-Targets (all-source)} which is a more challenging large-scale setting as source images can come from all the ImageNet classes except the target class. For consistency and direct comparison, the ten target classes are same as in \cite{Inkawhich2020Transferable}.

\noindent $-$\textit{10-Targets (subset-source):} Following  \cite{Inkawhich2020Transferable}, for each target class, $450$ source samples belonging to remaining $9$ classes (except target class) become inputs to $\gen_{\btheta}$ to be transferred to the selected target. 

\noindent $-$\textit{10-Targets (all-source):} For each target class, samples of all $999$ source classes (except the target class) in ImageNet val. set are considered i.e., for each target class, $49{,}950$ samples of $999$ classes become inputs to $\gen_{\btheta}$.

\begin{table}[!t]
	\centering\small
		\setlength{\tabcolsep}{2pt}
		\scalebox{0.80}[0.80]{
		\begin{tabular}{ @{\extracolsep{\fill}} L{0.5cm}|l|ccccc}
				\toprule
			\rowcolor{Gray} 
			& & \multicolumn{5}{c}{Naturally Trained (IN) Models} \\
			\cline{3-7}
			\rowcolor{Gray} 
			\multirow{-3}{*}{Src.} &\multirow{-3}{*}{Attack}& {VGG19$_{BN}$} & {Dense121} &  {ResNet50}& {ResNet152} & {WRN-50-2}\\
			\midrule
			\parbox[t]{2mm}{\multirow{6}{*}{\rotatebox[origin=c]{90}{VGG19$_{BN}$}}}   &PGD \cite{madry2018towards} &95.67$^*$ &0.31&0.30&0.20&0.25\\
			& MIM \cite{dong2018boosting}&99.91$^*$&0.92&0.68&0.36&0.47\\
			& DIM \cite{xie2019improving}&99.38$^*$&3.10&2.08&1.02&1.29\\
			& DIM-TI \cite{dong2019evading}&89.71$^*$&1.08&0.66&0.42&0.45\\
			& Po-TRIP \cite{li2020towards} & 99.40$^*$ & 4.61 & 3.21 & 1.78& 2.01 \\
			& GAP \cite{poursaeed2018generative}&98.23$^*$&16.19&15.83&5.89&7.78\\
			& CDA \cite{naseer2019cross}& 98.30$^*$ &16.26&16.22&5.73&8.35\\
			\cline{2-7}
			& \textbf{\texttt{Ours-P}} &97.38$^*$&45.53&42.90&26.72&31.00\\
			& \textbf{\texttt{Ours}} &98.54$^*$&\textbf{45.77}&\textbf{45.87}&\textbf{27.18}&\textbf{32.63}\\
			\midrule
\cls{V$_{ens}$}& \textbf{\texttt{Ours}}&97.34$^*$&71.41&71.68&50.78&48.03 \\
			\midrule
			\parbox[t]{2mm}{\multirow{6}{*}{\rotatebox[origin=c]{90}{Dense121}}}&PGD \cite{madry2018towards}&1.28&97.40$^*$&1.78&1.01&1.37\\
			& MIM \cite{dong2018boosting}&1.85&99.90$^*$&2.71&1.68&1.88\\
			& DIM \cite{xie2019improving}&7.31&98.81$^*$&9.06&5.78&6.29\\
			& DIM-TI \cite{dong2019evading}&0.91&88.59$^*$&1.18&0.77&0.86\\
			& Po-TRIP \cite{li2020towards} & 8.10 & 99.00$^*$&11.21 & 7.83 & 8.50\\
			& GAP \cite{poursaeed2018generative}&39.01&97.30$^*$&47.85&39.25&34.79\\
			& CDA \cite{naseer2019cross}& 42.77&97.22$^*$&54.28&44.11&46.01\\
			\cline{2-7}
			& \textbf{\texttt{Ours-P}} &57.91&97.41$^*$&\textbf{71.35}&55.57&53.45\\
			& \textbf{\texttt{Ours}} & \textbf{58.90}&97.61$^*$&68.72&\textbf{57.11}&\textbf{56.80}\\
			\midrule
\cls{D$_{ens}$}& \textbf{\texttt{Ours}} &76.96&96.25$^*$&88.81&83.48&81.85\\
			\midrule
			\parbox[t]{2mm}{\multirow{6}{*}{\rotatebox[origin=c]{90}{ResNet50}}}&PGD \cite{madry2018towards}&0.92&1.38&93.74$^*$&1.86&1.89\\
			& MIM \cite{dong2018boosting}&1.58&3.37&98.76$^*$&3.39&3.17\\
			& DIM \cite{xie2019improving}&9.14&15.47&99.01$^*$&12.45&12.61\\
			& DIM-TI \cite{dong2019evading}&0.79&2.12&88.91$^*$&1.47&1.45\\
			& Po-TRIP \cite{li2020towards} & 12.01 & 19.43 & 99.22$^*$ & 14.41 & 15.10\\
			& GAP \cite{poursaeed2018generative} & 58.47&71.72&96.81$^*$&64.89&61.82\\
			& CDA \cite{naseer2019cross} & 64.58 & 73.57 & 96.30$^*$ & 70.30 & 69.27  \\
			\cline{2-7}
            & \textbf{\texttt{Ours-P}} &73.09&\textbf{84.76}&96.63$^*$&76.27&75.92\\
			& \textbf{\texttt{Ours}} & \textbf{78.15} & 81.64 & 97.02$^*$ & \textbf{80.56} & \textbf{78.25}   \\
			\midrule
						  \cls{R$_{ens}$}& \textbf{\texttt{Ours}} & 90.43&94.39&96.67$^*$&95.48$^*$&92.63\\
			\bottomrule
	\end{tabular}}
	\caption{\emph{\textbf{Target Transferability:}} $\{$10-Targets (all-source)$\}$ Top-1 target accuracy (\%) averaged across 10 targets with 49.95K ImageNet val. samples. Perturbation budget: $l_\infty \le 16$. Our method outperforms previous instance-specific as well as generative approaches by a large margin. '*' indicates white-box attack. \textbf{\texttt{Ours-P}} represents \textsc{TTP} trained on Paintings.}
	\label{tab: undefended_imagenet_val}
\end{table}

\begin{table}[!t]
	\centering\small
		\setlength{\tabcolsep}{7pt}
		\scalebox{0.8}[0.8]{
		\begin{tabular} {@{\extracolsep{\fill}}  L{1.5cm}|l|c c c}
				\toprule
			\rowcolor{Gray} 
			Src.&Attack&VGG19$_{BN}$ & Dense121 & ResNet50\\
			\midrule
			\parbox[t]{2mm}{\multirow{5}{*}{\rotatebox[origin=c]{0}{VGG19$_{BN}$}}}& AA \cite{inkawhich2019feature} & --&0.8&0.6\\
			 & FDA-fd \cite{Inkawhich2020Transferable} &-- & 3.0&2.1\\
			 & FDA$^N$ \cite{inkawhich2020perturbing} & -- & 6.0 & 5.4 \\ 
			& CDA \cite{naseer2019cross}&--&17.82&17.09 \\
			\cline{2-5}
			& \textbf{\texttt{Ours-P}}&--&\textbf{48.56}&44.47 \\
			& \textbf{\texttt{Ours}}& --&48.29&\textbf{47.07} \\
			\midrule
			\parbox[t]{2mm}{\multirow{5}{*}{\rotatebox[origin=c]{0}{Dense121}}}  	& AA \cite{inkawhich2019feature} & 0.0 & -- & 0.0 \\
			& FDA-fd \cite{Inkawhich2020Transferable}&34.0 & --& 34.0\\
			& FDA$^N$ \cite{inkawhich2020perturbing}  & 42.0 & -- & 48.3 \\
			& CDA \cite{naseer2019cross}&44.84&--&53.73 \\
			\cline{2-5}
			& \textbf{\texttt{Ours-P}}&59.81&&\textbf{71.32} \\
			& \textbf{\texttt{Ours}}&\textbf{61.75}&--&69.60 \\
			\midrule
			\parbox[t]{2mm}{\multirow{5}{*}{\rotatebox[origin=c]{0}{ResNet50}}} & AA \cite{inkawhich2019feature} & 1.1 & 2.0 & -- \\ 	
			& FDA-fd \cite{Inkawhich2020Transferable}&16.0&21.0&-- \\
			& FDA$^N$ \cite{inkawhich2020perturbing} &32.1&48.3&--\\
			& CDA \cite{naseer2019cross} & 68.55&75.68&-- \\
			\cline{2-5}
			& \textbf{\texttt{Ours-P}}&75.18 & \textbf{85.71} \\
			& \textbf{\texttt{Ours}}&\textbf{79.04}&84.42&-- \\

			\bottomrule
	\end{tabular}}
	\caption{\emph{\textbf{Target Transferability:}}$\{$10-Targets (sub-source)$\}$ Top-1 accuracy (\%) across 10 targets. Our method shows significant improvements in trasfering target perturbations compared to generative as well as feature based instance-specific method \cite{Inkawhich2020Transferable, inkawhich2020perturbing}. Perturbation budget:$l_\infty \le 16$. Only black-box attack results are shown. \textbf{\texttt{Ours-P}} represents \textsc{TTP} trained on Paintings.} 
	\label{tab: undefended_imagenet_val_subset}

\end{table}

\noindent\textbf{100-Targets (all-source):}
We divide ImageNet 1k classes into 100 mutually exclusive sets. Each set contains 10 classes. We randomly sample 1 target from each set to create 100 targets (see \ref{sec: sup_target_names} for more details). Generators are trained against these targets and evaluated on ImageNet val. set in 100-Targets (all-source) setting with the same protocol as described for 10-Targets (all-source). 

\subsection{Attack Protocols and Results}

We evaluate black-box target transferability in the following scenarios. \textbf{(a)} \emph{\underline{Unknown Target Model:}} Attacker has access to a pretrained discriminator trained on labeled data but has no knowledge about the architecture of the target model. 
\textbf{(b)}  \emph{\underline{Unknown Decision Space:}} Attacker has access to the pre-trained discriminator trained on unlabeled data in an unsupervised manner but does not know about the architecture and the class-boundary information learned by the target model.
\textbf{(c)} \emph{\underline{Unknown Defense:}} Attacker is unaware of the type of defense deployed at the target model, or if any defense is applied at all, e.g., the defense can be an input processing approach or a robust training mechanism such as adversarial training.

\subsubsection{Unknown Target Model}
\noindent\textbf{Natural Training:} We evaluate naturally trained ImageNet models and show strong empirical results in Tables \ref{tab: undefended_imagenet_val}, \ref{tab: undefended_imagenet_val_subset} \& \ref{tab: undefended_imagenet_val_100_targets} demonstrating that generative methods are far superior than sample-specific targeted attacks based on boundary information \cite{li2020towards, dong2018boosting, xie2019improving} or feature exploitation \cite{inkawhich2019feature, Inkawhich2020Transferable, inkawhich2020perturbing}. Our approach has significantly higher target transferability rates than previous generative methods \cite{naseer2019cross, poursaeed2018generative}. To highlight an example from Table~\ref{tab: undefended_imagenet_val_subset}, our method achieves $47.07\%$ transferability from VGG19$_{BN}$ to ResNet50 which is $175\%$ and $771\%$ better than the previous best generative \cite{naseer2019cross} and sample-specific \cite{inkawhich2020perturbing} target attacks, respectively.

\noindent\textbf{Ensemble Effect:} We also train generators with our algorithm on the ensembles of same-family discriminators. Specifically, we define the following ensembles: 
    \cls{V$_{ens}$:VGG\{11,13,16,19\}$_{BN}$},\cls{R$_{ens}$:ResNet\{18,50,101, 152\}}, and \cls{D$_{ens}$:DenseNet\{121,161,169,201\}}.

\begin{figure*}[t]
\centering
\begin{minipage}{0.66\textwidth}

    \begin{minipage}{.33\textwidth}
  	\centering
    \includegraphics[ width=\linewidth, height=2.8cm]{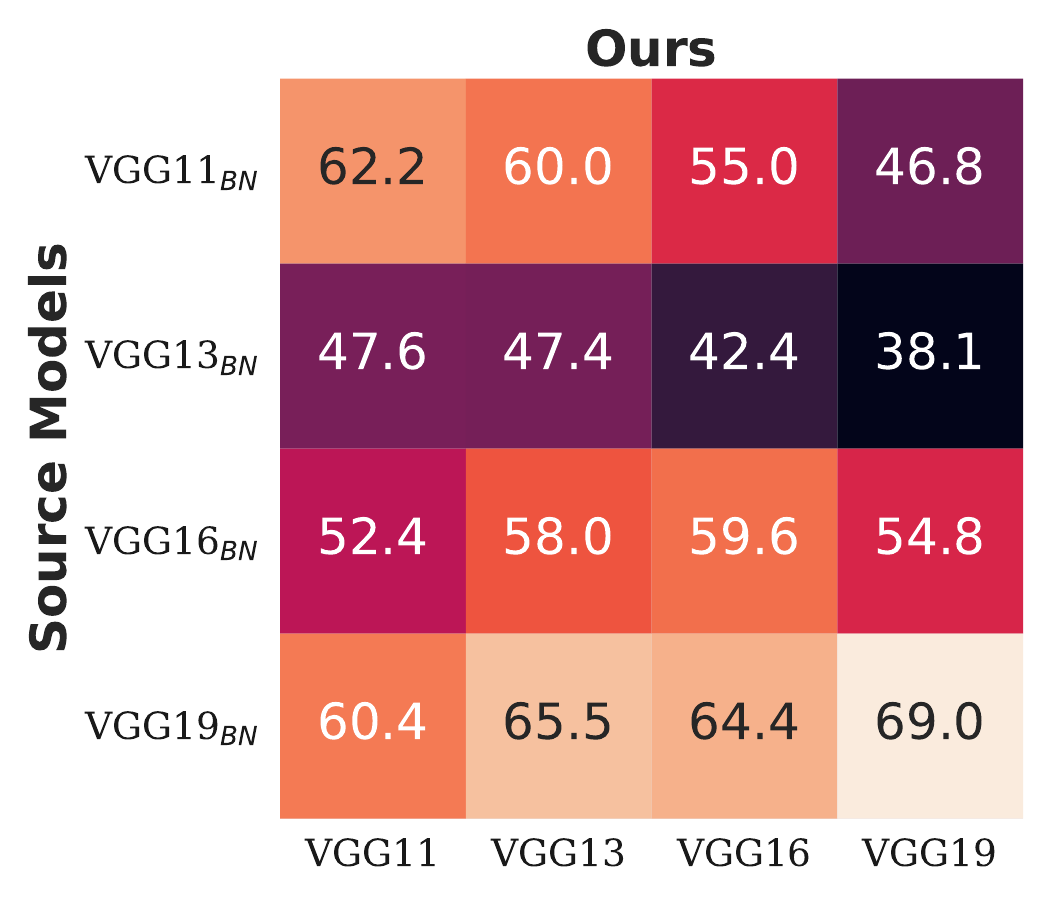}\
  \end{minipage}
    \begin{minipage}{.315\textwidth}
  	\centering
    \includegraphics[width=\linewidth, height=2.8cm]{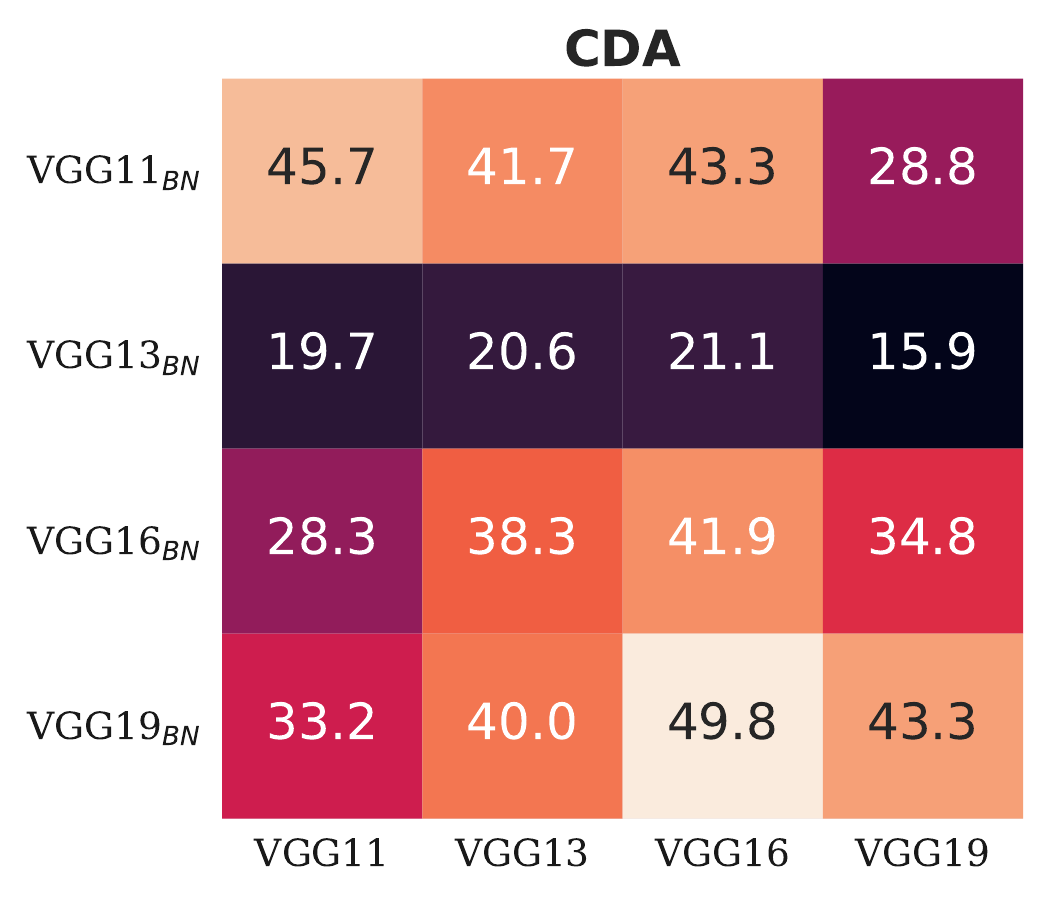}\
  \end{minipage}
    \begin{minipage}{.315\textwidth}
  	\centering
    \includegraphics[ width=\linewidth, height=2.8cm]{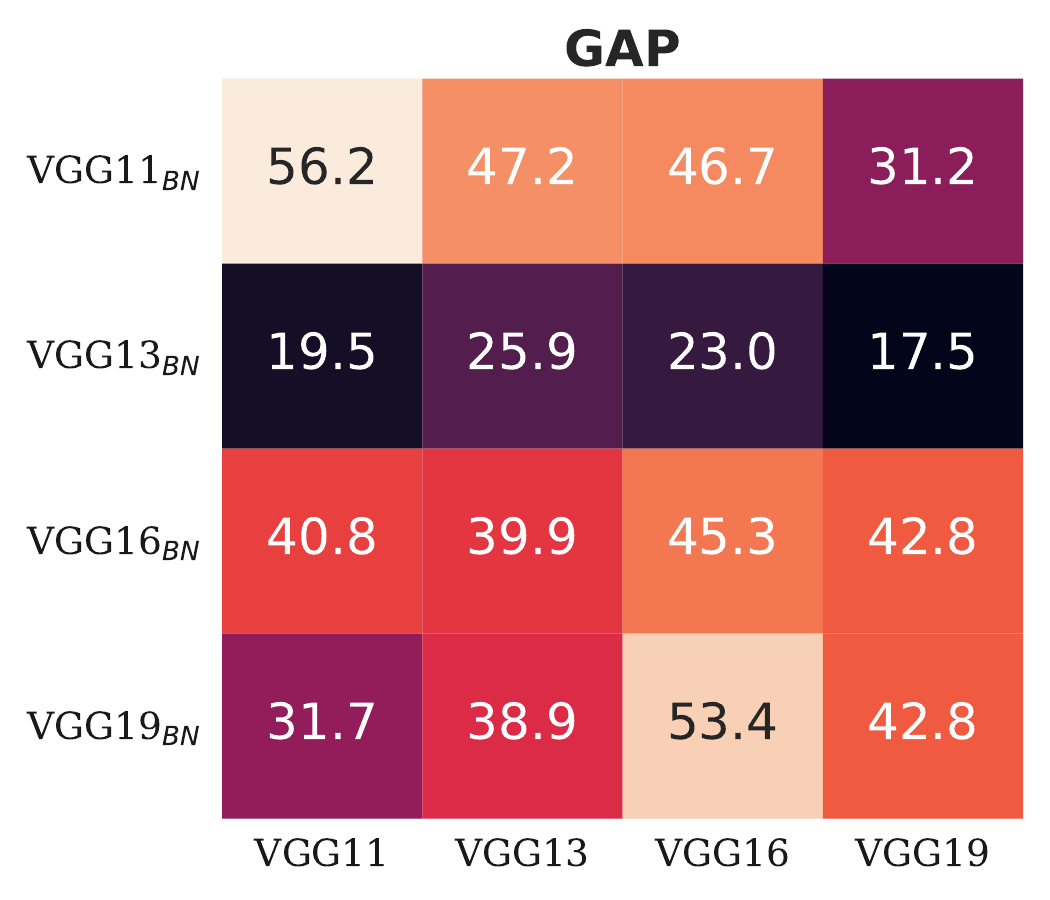}\
  \end{minipage}
  \\
    \begin{minipage}{.33\textwidth}
  	\centering
    \includegraphics[ width=\linewidth, keepaspectratio]{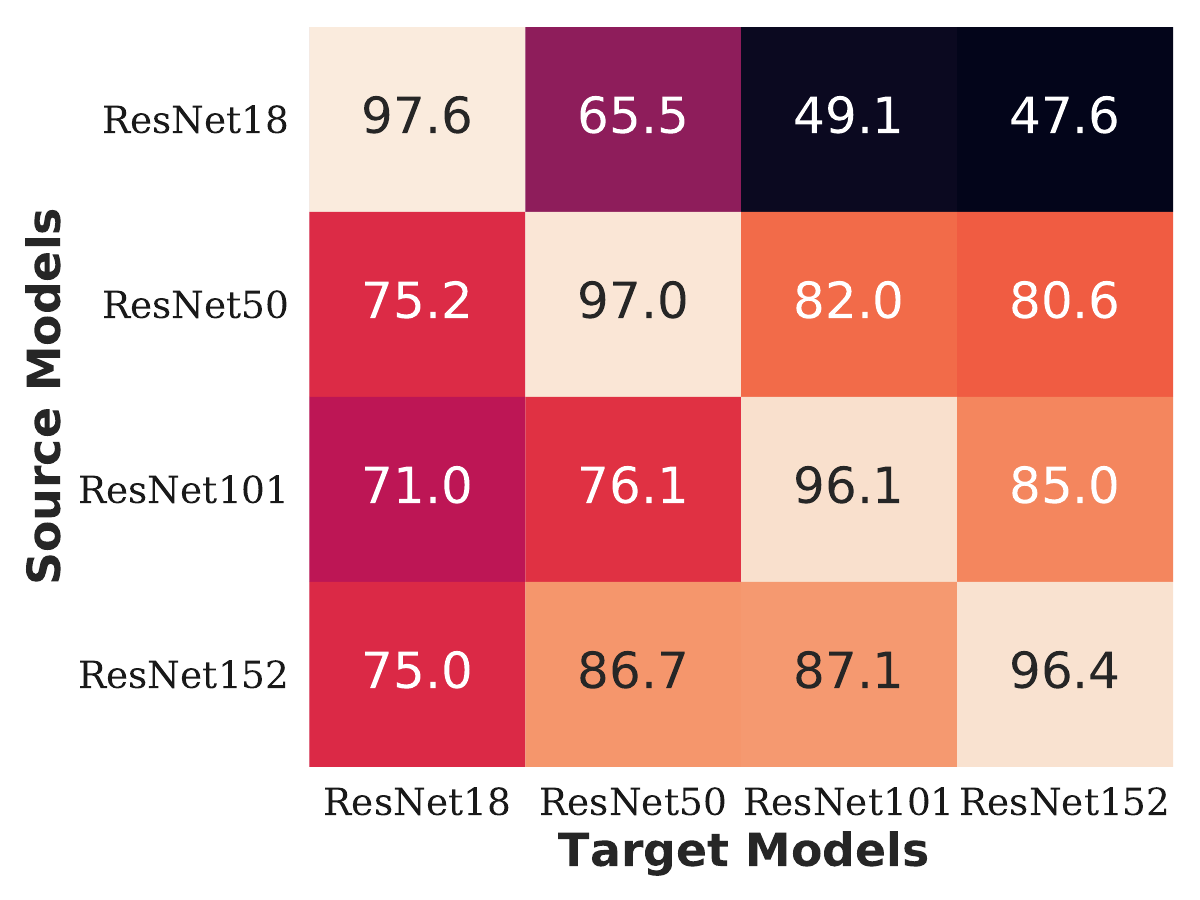}\
  \end{minipage}
    \begin{minipage}{.315\textwidth}
  	\centering
    \includegraphics[width=\linewidth, keepaspectratio]{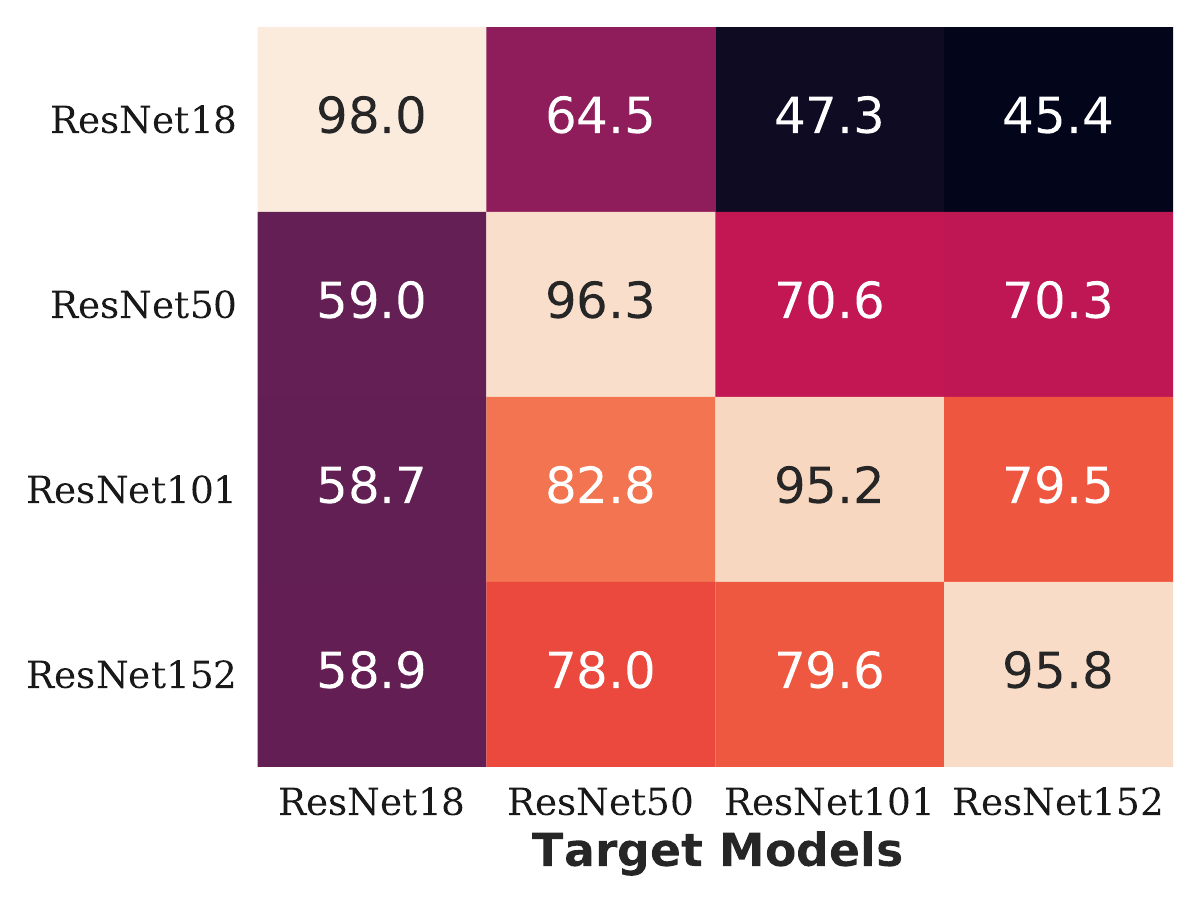}\
  \end{minipage}
    \begin{minipage}{.315\textwidth}
  	\centering
    \includegraphics[ width=\linewidth, keepaspectratio]{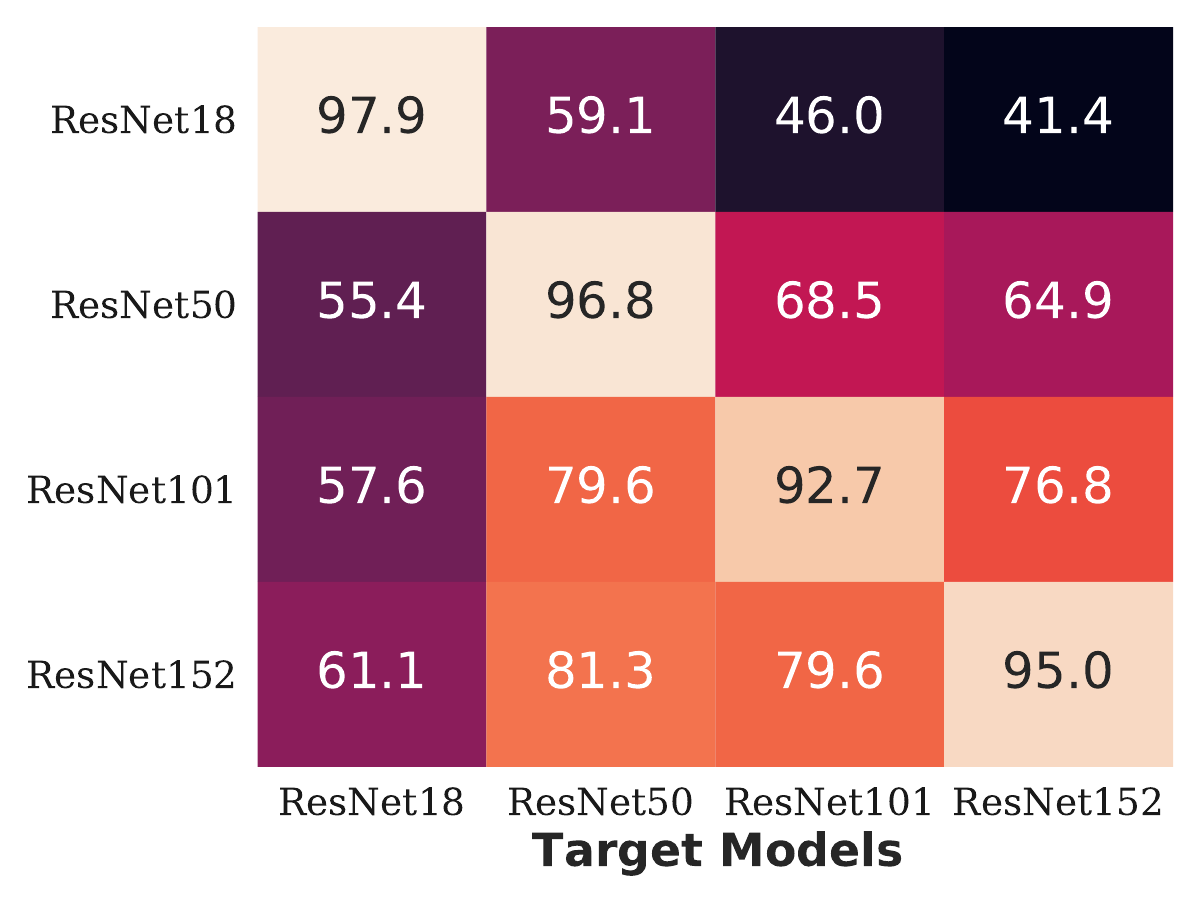}\
  \end{minipage}
\end{minipage}
\hfill
\begin{minipage}{.33\textwidth}
\captionof{figure}{\emph{\textbf{Within Family Target Transferability:}} $\{$10-Targets (all source) settings$\}$ These results indicate that our approach boosts target transferability within different models of the same family with or without batch-norm and favorably beats the previous generative approaches (GAP \cite{poursaeed2018generative}, CDA \cite{naseer2019cross})  by a large margin. Each value is averaged across 10 targets (Sec.~\ref{sec: experiments}) with 49.95k ImageNet val. samples for each target. Perturbation budget is set to $l_\infty =16$.}
\label{fig: heatmps}
\end{minipage}
\end{figure*}

\begin{figure*}[t]
\centering
\begin{minipage}{.66\textwidth}
      \begin{minipage}{.32\textwidth}
  	\centering
    \includegraphics[ width=\linewidth, height=3.1cm]{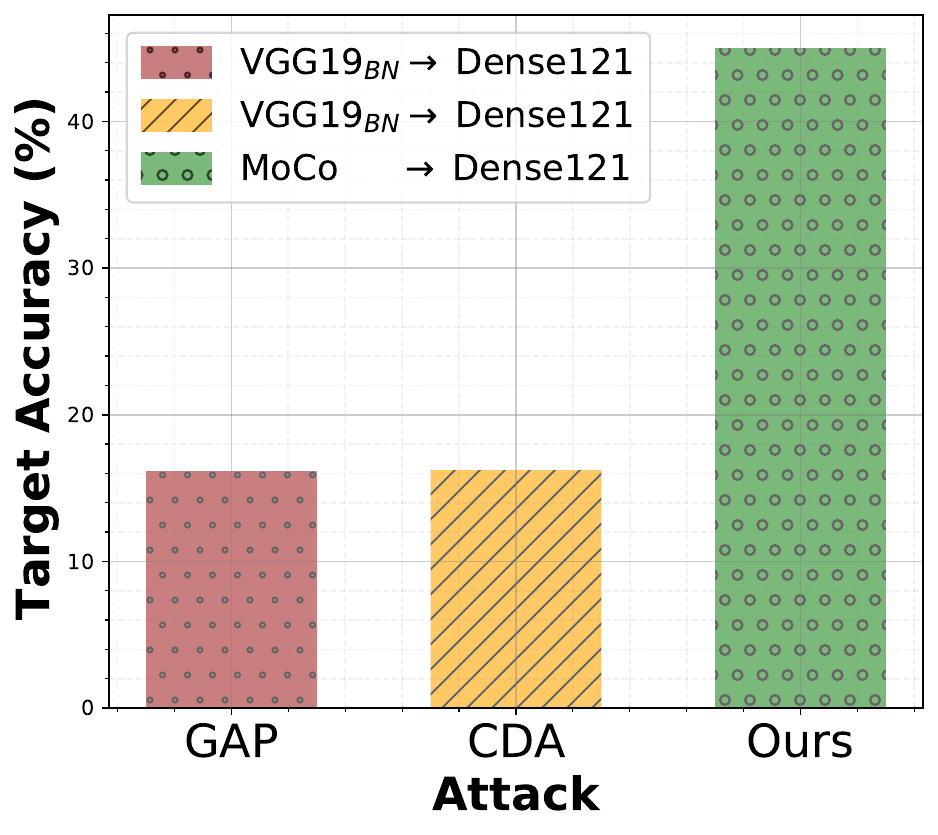}\
  \end{minipage}
    \begin{minipage}{.32\textwidth}
  	\centering
    \includegraphics[width=\linewidth, height=3.1cm]{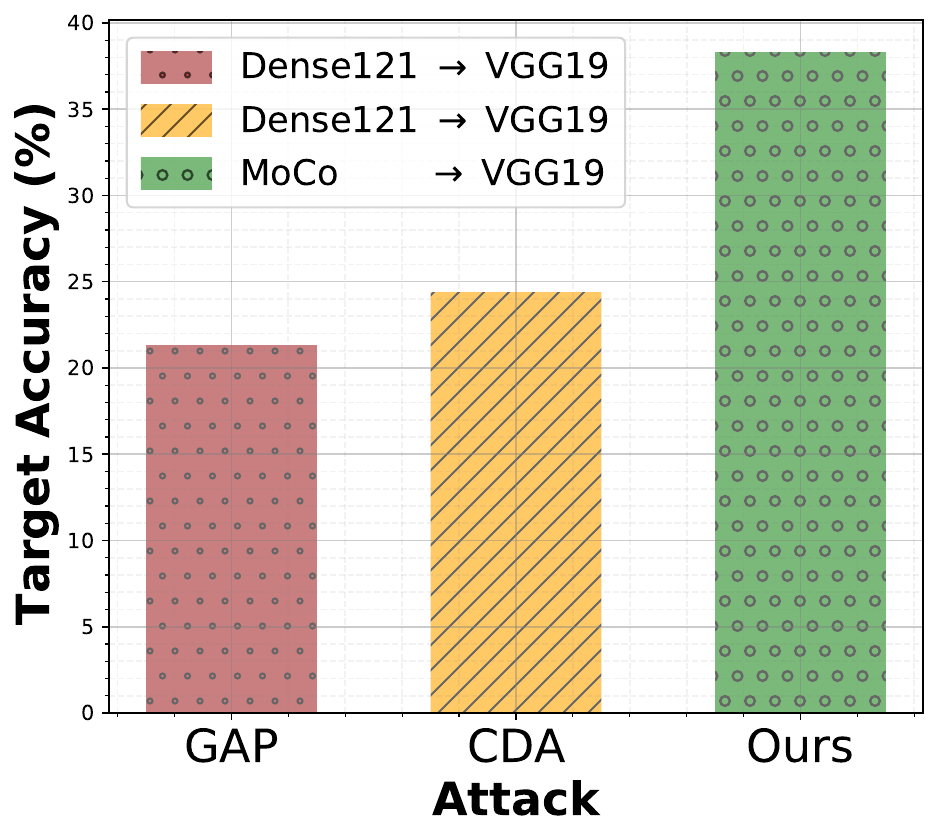}\
  \end{minipage}
    \begin{minipage}{.32\textwidth}
  	\centering
    \includegraphics[ width=\linewidth, height=3.1cm]{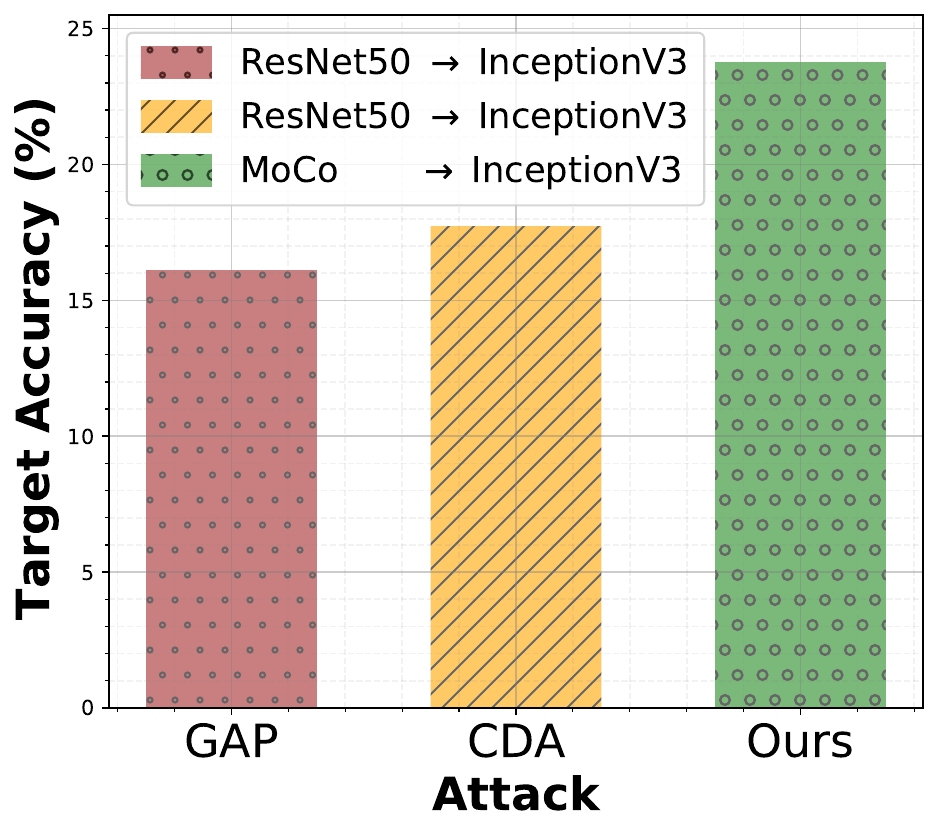}\
  \end{minipage}
\end{minipage}
 \hfill
\begin{minipage}{.33\textwidth}
\captionof{figure}{\emph{\textbf{Target Transferability of Unsupervised Features:}} $\{$10-Targets (all-source) settings$\}$. Our approach when applied to unsupervised features, MoCo \cite{he2020momentum}, surpasses GAP \cite{poursaeed2018generative} and CDA \cite{naseer2019cross} that are dependent on classification layer by design. Perturbation budget is $l_\infty=16$.} 
\label{fig: supervised_vs_unsupervised}
\end{minipage}
\end{figure*}

\noindent The purpose of such ensembles is to understand if the combination of weak individual models from the same family can provide strong learning for the target distributions. From Table~\ref{tab: undefended_imagenet_val}, we observe that modeling target distribution from an ensemble provides significantly better tranferability than any individual discriminator (see \ref{sec: sup_per_class_analysis} for more analysis). This signifies that an attacker can use multiple variants of the same network to boost the attack.

\noindent\textbf{Target Transferability and Model Disparity:} 
We note that within a specific family, transferring targeted perturbations from a smaller model to a larger one (\eg ResNet18 $\rightarrow$ ResNet152 or VGG11$_{BN}$ $\rightarrow$ VGG19) is difficult as we increase the size discrepancy. Interestingly, this trend remains the same even from larger to smaller models i.e., the attack strength will increase with the disparity between models rather than only depending upon the strength of target model. For example, target transferability  ResNet152 $\rightarrow$ ResNet50 is higher than ResNet152 $\rightarrow$ ResNet18 even though ResNet18 is weaker than ResNet50 (Fig.~\ref{fig: heatmps}). Similar behaviour can be observed within cross-family models \ie, target transferability from ResNet50 to Dense121 and vice versa is higher than VGG19$_{BN}$ as both models share skip connections (Table ~\ref{tab: undefended_imagenet_val}). See \ref{sec: sup_batchnorm} for vulnerability of models with and without batch-norm \cite{ioffe2015batch}.

\begin{figure*}[t]
\centering
\begin{minipage}{.66\textwidth}
      \begin{minipage}{.24\textwidth}
  	\centering
    \includegraphics[ width=\linewidth, keepaspectratio]{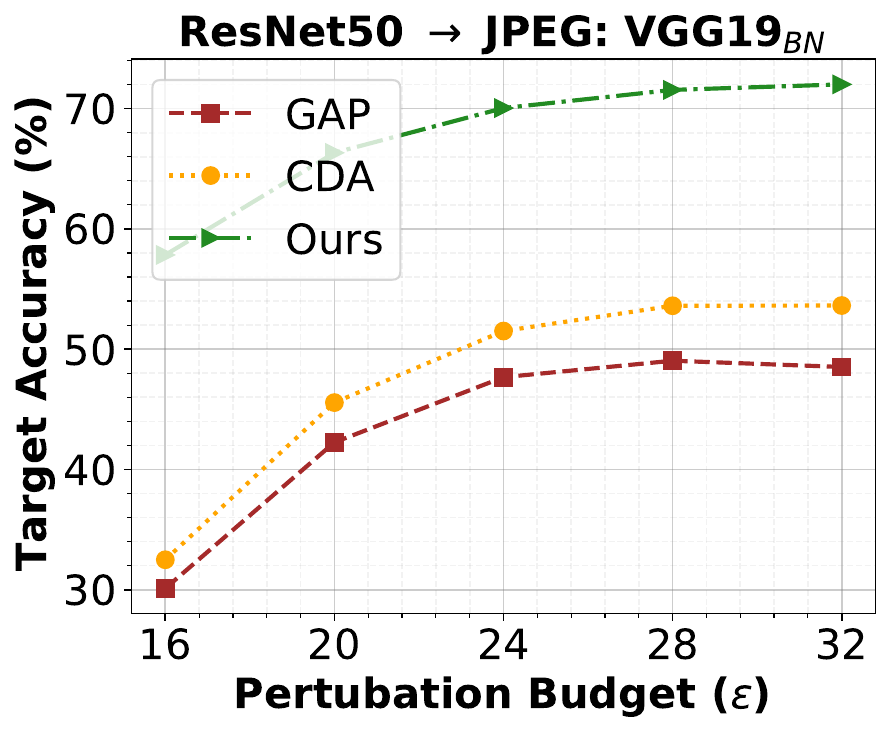}\
  \end{minipage}
    \begin{minipage}{.24\textwidth}
  	\centering
    \includegraphics[ width=\linewidth, keepaspectratio]{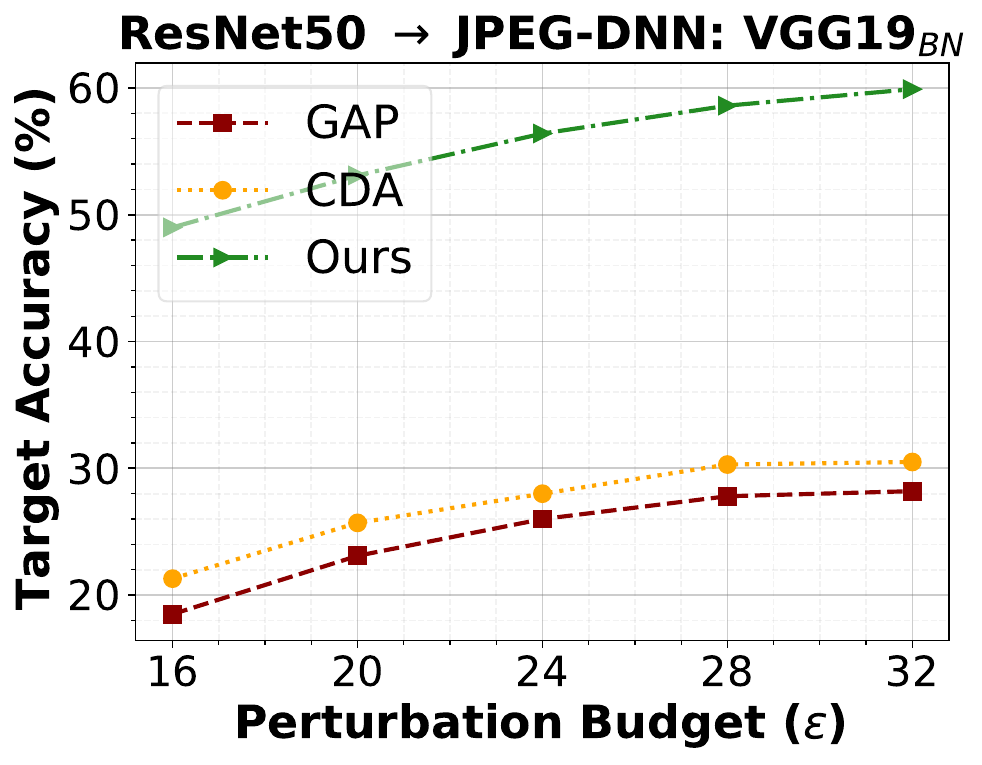}\
  \end{minipage}
    \begin{minipage}{.24\textwidth}
  	\centering
    \includegraphics[width=\linewidth, keepaspectratio]{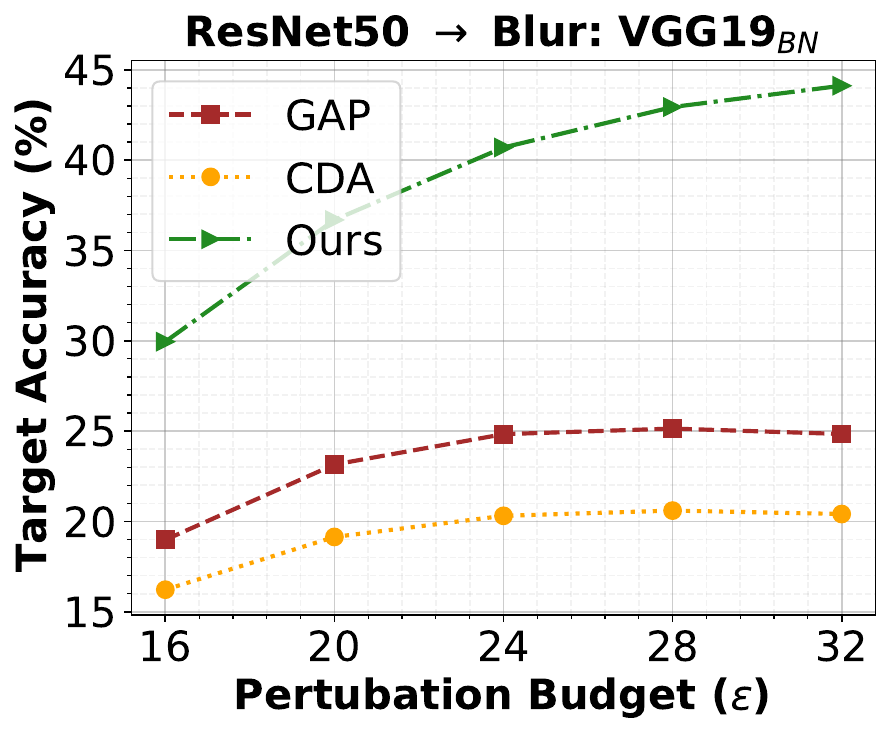}\
  \end{minipage}
    \begin{minipage}{.24\textwidth}
  	\centering
    \includegraphics[ width=\linewidth, keepaspectratio, trim=0 0 10 0mm]{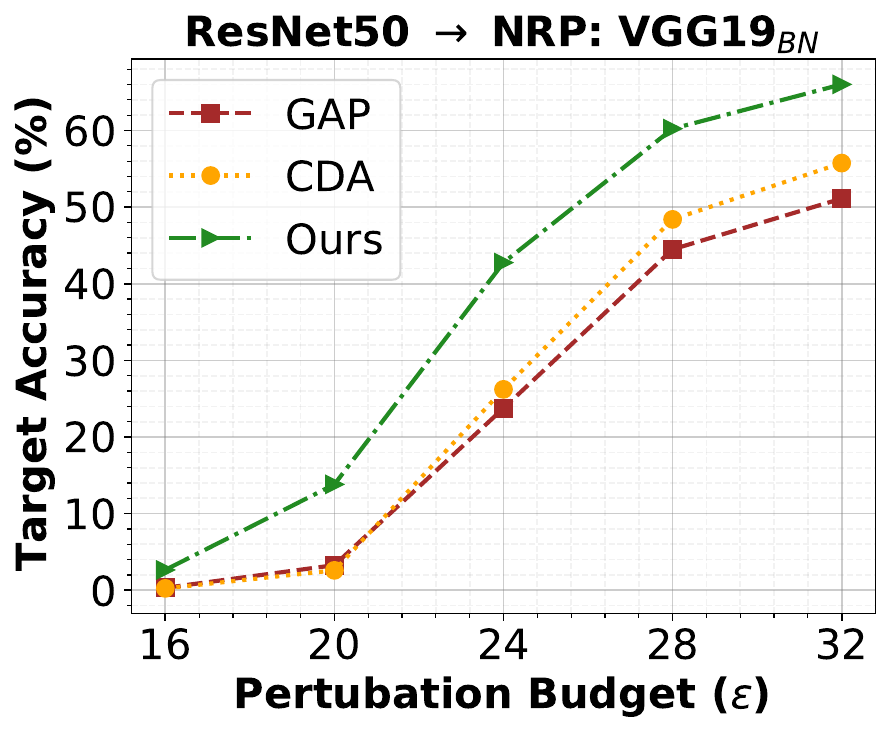}\
  \end{minipage}
  \\
  \begin{minipage}{.24\textwidth}
  	\centering
    \includegraphics[ width=\linewidth, keepaspectratio]{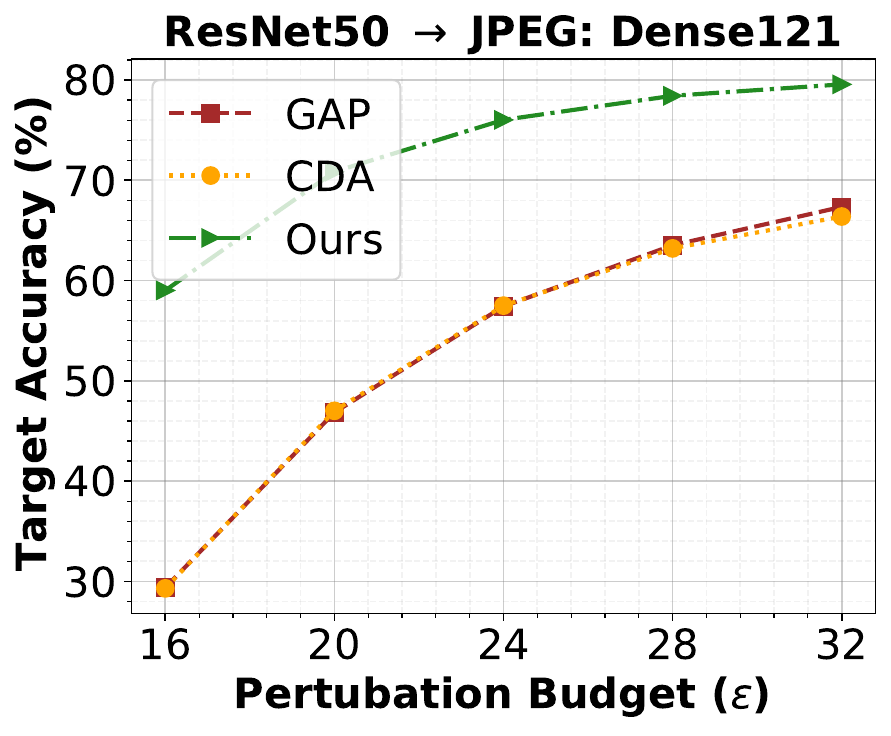}\
  \end{minipage}
   \begin{minipage}{.24\textwidth}
  	\centering
    \includegraphics[ width=\linewidth, keepaspectratio]{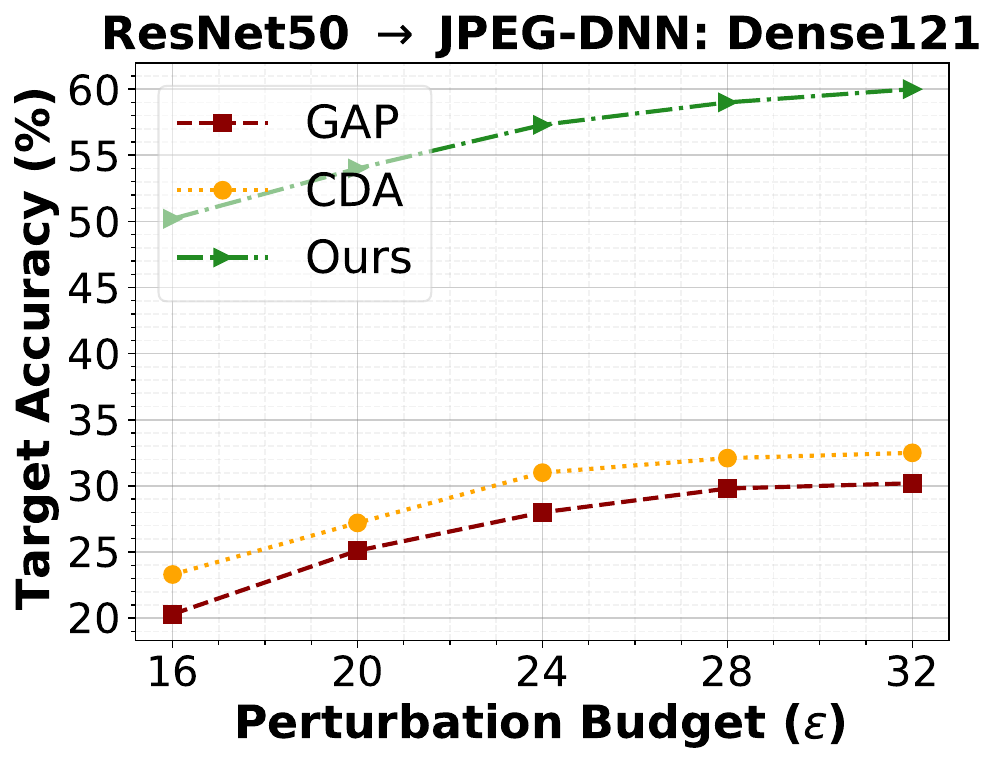}\
  \end{minipage}
    \begin{minipage}{.24\textwidth}
  	\centering
    \includegraphics[width=\linewidth, keepaspectratio]{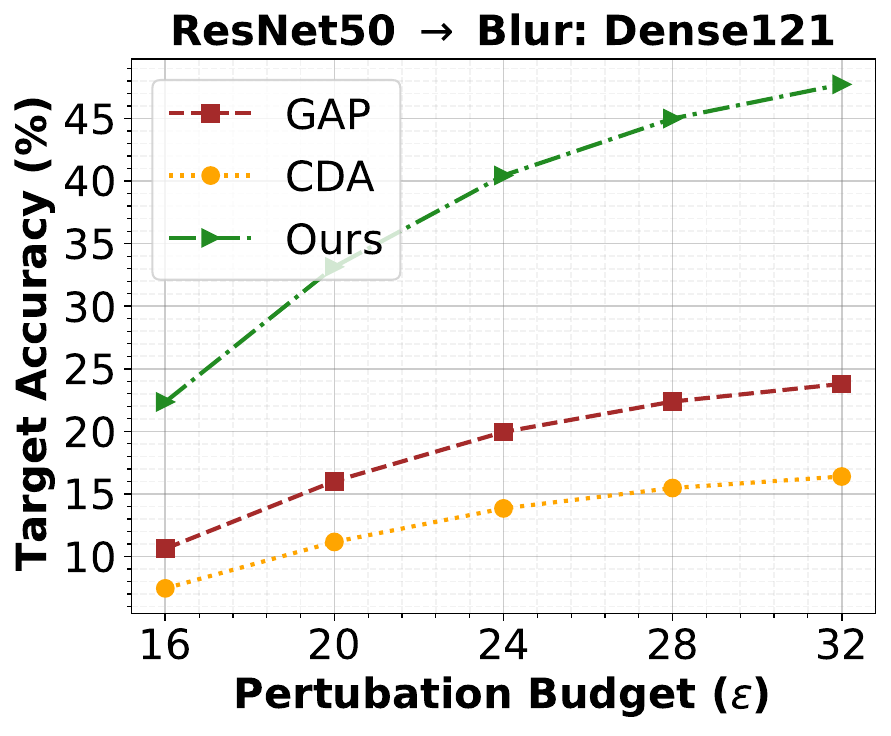}\
  \end{minipage}
    \begin{minipage}{.24\textwidth}
  	\centering
    \includegraphics[ width=\linewidth, keepaspectratio, trim=0 0 10 0mm]{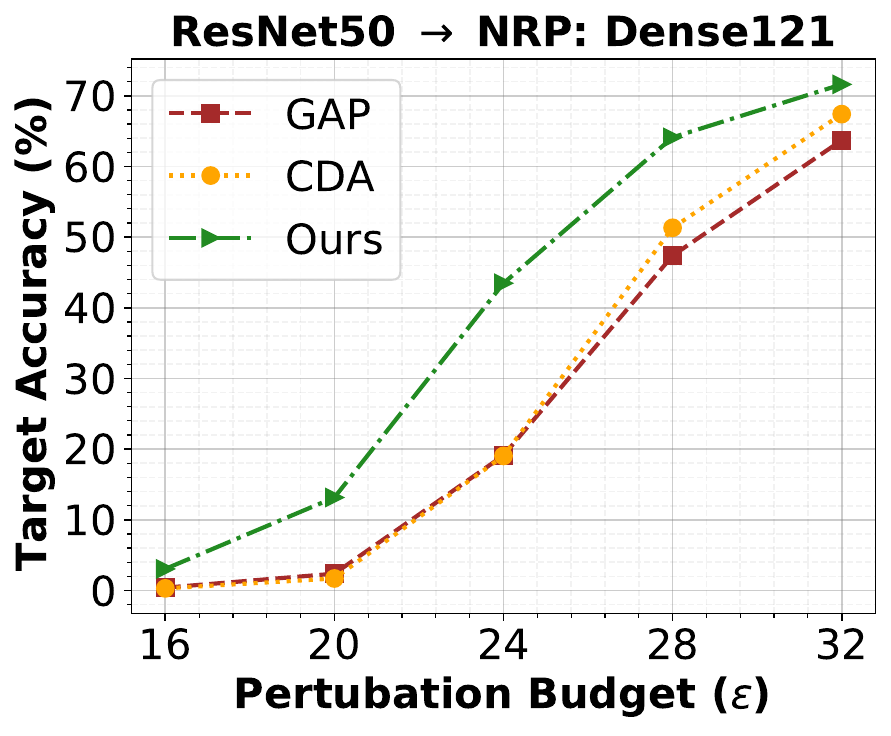}\
  \end{minipage}
\end{minipage}
 \begin{minipage}{0.33\textwidth}
\captionof{figure}{\emph{\textbf{Target Transferability against Input Processing Defenses:}} $\{$10-Targets (all-source) settings$\}$ Input processing including NRP \cite{Naseer_2020_CVPR} are broken under targeted black-box attacks. Our method outperforms GAP \cite{poursaeed2018generative} and CDA \cite{naseer2019cross} on all the considered defenses including JPEG, JPEG-DNN \cite{liu2018feature}, Median Blur and NRP \cite{Naseer_2020_CVPR}. Each point is an averaged across 10 targets (Sec.~\ref{sec: experiments}) with 49.95k ImageNet val. samples for each target. Generators are trained against ResNet50.}
\label{fig: input_processsing_defenses}
\end{minipage}
 
\end{figure*}

\vspace{-1em}
\subsubsection{Unknown Decision Space}
Here we investigate the question, ``\textit{Can unsupervised features provide targeted adversarial perturbations?}" A unique property of our proposed approach is that it can be applied to feature space without any class boundary information to achieve target adversarial direction. This allows an attacker to benefit from recently proposed unsupervised feature learning methods \cite{he2020momentum, chen2020simple}. Rather than using a discriminator trained on large scale labelled data, attack can be learned and launched from features of a discriminator trained purely in an unsupervised fashion. Therefore, our attack can eliminate the cost of label annotations. Results in Fig.~\ref{fig: supervised_vs_unsupervised} demonstrate that our method learned from unsupervised features, MoCo \cite{he2020momentum}, not only provides target transferability but surpasses the previous generative methods which are dependent on the discriminator trained on labelled data.

\begin{table}[!t]
	\centering\small
		\setlength{\tabcolsep}{4pt}
		\scalebox{0.85}{
		\begin{tabular}{ @{\extracolsep{\fill}} l| c c c c c}
				\toprule
            \rowcolor{Gray} 
			Attack & {VGG19$_{BN}$} & {Dense121} & {ResNet-152} & {WRN-50-2} & SIN \cite{geirhos2018imagenettrained}\\
            \midrule
		  GAP \cite{poursaeed2018generative}&47.87&58.10&54.72&49.65&7.1\\
		 CDA \cite{naseer2019cross} &53.41&60.34&57.67&51.23&7.6 \\
		 \textbf{\texttt{Ours}} & \textbf{69.55}&\textbf{77.48}&\textbf{75.74}&\textbf{74.61}&\textbf{31.0}  \\
			\bottomrule
	\end{tabular}}
	\caption{\emph{\textbf{Target Transferability:}} $\{$100-Targets (all-source)$\}$ Top-1 target accuracy (\%) averaged across 100 targets with 49.95K ImageNet val. samples per target . Generators are trained against ResNet50. Perturbation budget is $l_\infty \le 16$. }
	\label{tab: undefended_imagenet_val_100_targets}
\end{table}

\begin{table}[t]\setlength{\tabcolsep}{5pt}
  \centering 
  \scalebox{0.75}{
  \begin{tabular}{llccccccc} 
    \toprule
    \rowcolor{Gray} 
     $\epsilon$ & Attack   & Augmix    &  \multicolumn{2}{c}{Stylized \cite{geirhos2018imagenettrained}}   & \multicolumn{4}{c}{Adversarial \cite{salman2020adversarially}}  \\
     \cmidrule(lr){4-5}  \cmidrule(lr){6-9}
      &  & \cite{hendrycks2020augmix} &SIN-IN & SIN  &   \multicolumn{2}{c}{$l_\infty$ }&\multicolumn{2}{c}{$l_2$} \\
       \cmidrule(lr){6-7} \cmidrule(lr){8-9}
      && &       &  & $\epsilon{=}.5$&  $\epsilon{=}1$&$\epsilon{=}.1$&$\epsilon{=}.5$ \\   
    \midrule
    \multirow{4}{*}{16}  & GAP \cite{poursaeed2018generative} &51.57&76.92&12.96&1.88&0.34&23.41&0.92 \\
    &CDA \cite{naseer2019cross} & 59.79& 75.93&9.21&2.10&0.39&23.89&1.18\\
     &Ours& \textbf{73.09}&\textbf{87.40}&\textbf{30.17}&\textbf{4.63}&\textbf{0.56}&\textbf{45.40}&\textbf{1.99}\\
    \cline{2-9}
    &Ours$_{ens}$  &88.79&92.96&57.75&14.23&1.24&74.95&7.62\\
    \midrule
    \multirow{4}{*}{32}  & GAP \cite{poursaeed2018generative} & 54.86 & 81.15&28.07&26.32&6.36&59.04&16.53\\
    &CDA \cite{naseer2019cross} &63.18&76.81&19.65&27.60&6.74&57.54&16.07 \\
     &Ours& \textbf{78.66}&\textbf{91.27}&\textbf{41.52}&\textbf{46.82}&\textbf{16.35}&\textbf{75.97}&\textbf{30.94}\\
    \cline{2-9}
    &Ours$_{ens}$  &89.96 &94.15 &70.70 &70.22&34.21&90.42&58.25\\
    \bottomrule
  \end{tabular}}
  \caption{\emph{\textbf{Target Transferability:}} $\{$10-Targets (all source) settings$\}$ Top-1 (\%) target accuracy. Generators are trained against naturally trained ResNet50 or ResNet ensemble. Perturbation are then transferred to ResNet50 trained using different methods including Augmix \cite{hendrycks2020augmix},  Stylized  \cite{geirhos2018imagenettrained} or adversarial \cite{salman2020adversarially}. 
  }
  \label{tab: augmentation_vs_stylized_vs_adv}
  \vspace{-0.3em}
\end{table}

\subsubsection{Unknown Defense Mechanisms}
 \noindent\textbf{Input Processing as a Defense:} We evaluate robustness of different input processing based adversarial defense methods in Fig.~\ref{fig: input_processsing_defenses}. We consider the following four representative defenses: a) JPEG with compression quality set to 50\% \cite{das2018shield}, b)  DNN-Oriented JPEG compression \cite{liu2018feature}, c) Median Blur with window size set to 5$\times$5 \cite{naseer2019local}, and d) Neural representation purifier (NRP) \cite{Naseer_2020_CVPR} which is a state-of-the-art defense. Generators are trained against naturally trained ResNet50 and target perturbations are then transferred to VGG19$_{BN}$ and Dense121 which are protected by the input processing defenses. We observe (Fig.~\ref{fig: input_processsing_defenses}) that JPEG is the least effective method against target attacks while JEPG-DNN \cite{liu2018feature} performs relatively better than JPEG. Compared to JPEG, JPEG-DNN and Median blur, NRP shows better resistance to target attacks at $l_\infty \le 16$ but quickly breaks as perturbation is increased. Median blur shows more resistance than JPEG, JPEG-DNN and NRP at higher perturbation rates ($l_\infty \le 32$)\footnote{Blur defense causes large drop in clean accuracy (see \ref{sec: sup_clean_acc_vs_defenses}).}. Success rate of our method is much better than previous generative attacks \cite{poursaeed2018generative, naseer2019cross} even when the target model and the input processing remain unknown (Fig.~\ref{fig: input_processsing_defenses}).

 \noindent\textbf{Robust Training Mechanism:} Here we study the transferability of our approach against various robust training methods (augmented vs. stylized vs. adversarial) based defense strategies.  Augmentation based training can make the model robust to natural corruptions \cite{hendrycks2020augmix} while training on stylized ImageNet \cite{geirhos2018imagenettrained} improves shape bias and training on adversarial examples can improve robustness against adversarial attacks at the cost of computation, clean accuracy, and generalization to global changes \cite{ford2019adversarial}. We evaluate the vulnerability of these training methods in Table~\ref{tab: augmentation_vs_stylized_vs_adv}. Generators are trained against naturally trained ResNet50 or ResNet ensemble and adversarial perturbations are then transferred to ResNet50 trained using Augmix \cite{hendrycks2020augmix}, Stylized ImageNet (SIN) \cite{geirhos2018imagenettrained}, mixture of Stylized and natural ImageNet (SIN-IN) and adversarial examples \cite{salman2020adversarially}. Target transferability can easily be achieved against models trained on mixture (SIN-IN), however, the model trained on stylized images (SIN) shows higher resistance but remains vulnerable as our target attack (ensemble) achieves $\approx$71\% success at perturbation of $l_\infty=32$ (Table~\ref{tab: augmentation_vs_stylized_vs_adv}). Adversarially trained models using Madry's method \cite{madry2018towards} are more robust to target attacks.  
 
\begin{figure}[t]
\centering
\begin{minipage}{.23\textwidth}
    \includegraphics[width=\linewidth, height=3.2cm]{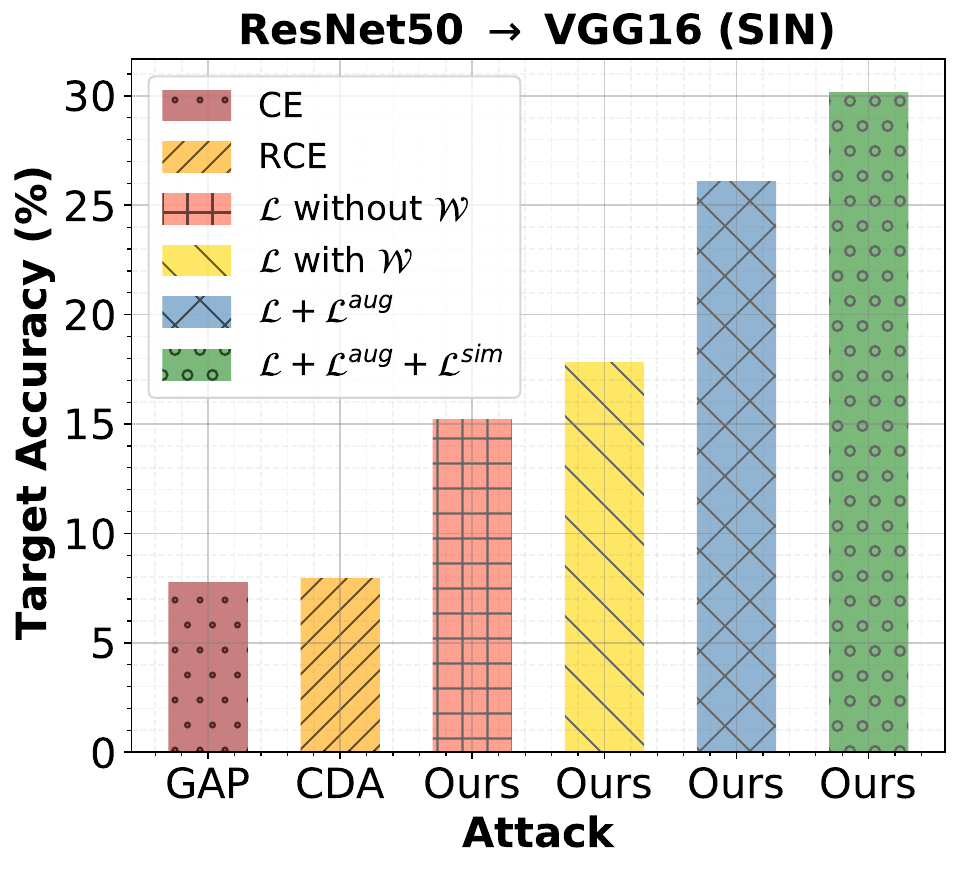}\
    \end{minipage}
\begin{minipage}{.23\textwidth}
\includegraphics[width=\linewidth, height=3.2cm]{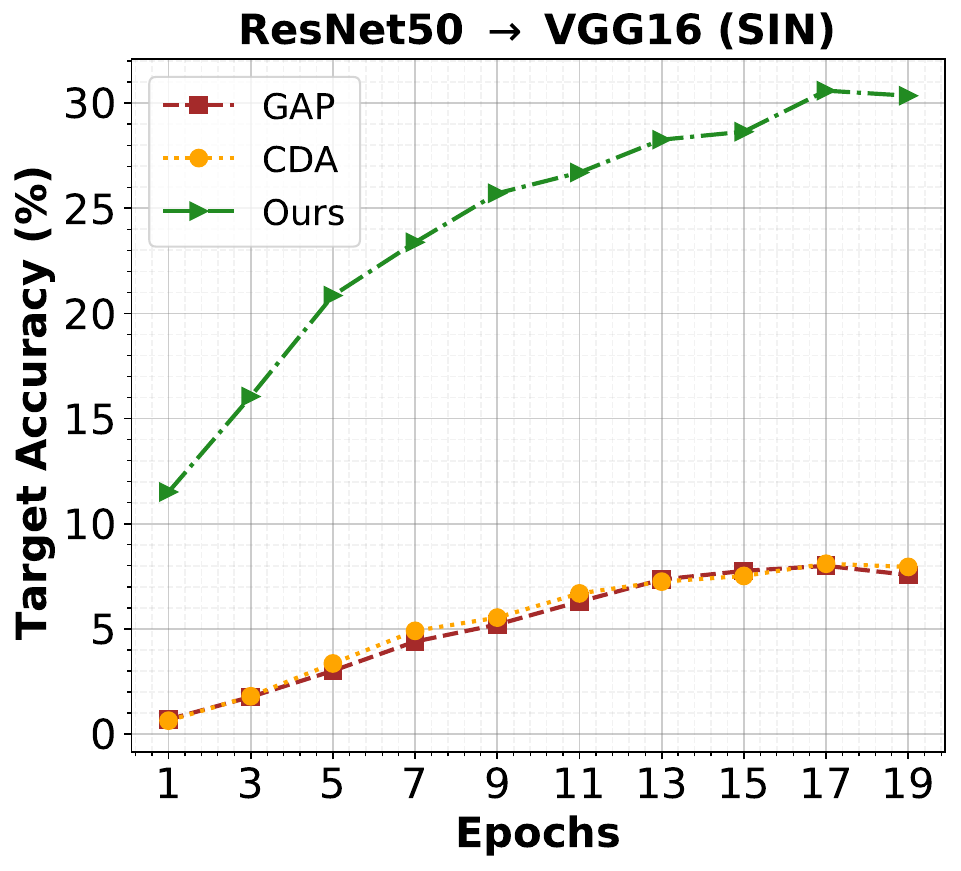}\
\end{minipage} \vspace{-0.5em}
  \caption{\emph{\textbf{Ablation:}} We dissect effect of each component of our method including novel losses, augmentation, smooth projection and epochs.  Results are presented with 10-Target (all source) settings. Perturbation budget is set to $l_\infty=16$. 
  }
\label{fig: loss_ablation}
\end{figure}

\subsection{Ablative Analysis}
In order to understand the effect of each component of our approach, we present an ablative study in Fig. \ref{fig: loss_ablation}. Target perturbations are transferred from ResNet50 to VGG16 (SIN) trained on stylized ImageNet which is a much harder task than transferring to naturally trained VGG16. We observe that training \ours on only distribution matching loss (Eq.~\ref{eq: loss_on_source}) increases the transferability by more than 100\% in comparison to GAP \cite{poursaeed2018generative} (with cross-entropy) or CDA \cite{naseer2019cross} (with relativistic cross-entropy). Adding smoothing operator $\mathcal{W}$ enhances the efficiency of \ours. $\mathcal{W}$ is a differentiable Gaussian kernel with size 3$\times$3. We then noticed a significant jump in transferability when augmentations are introduced and \ours is trained using both distribution matching losses (Eq. \ref{eq: loss_on_source} \& \ref{eq: loss_on_aug}) which is further complemented by neighbor similarity loss (Eq. \ref{eq: loss_sim}). Our generator trained for only one epoch outperforms GAP and CDA trained for 20 epochs (Fig. \ref{fig: loss_ablation}) which highlights our rapid convergence rate. 

\vspace{-0.2em}
\section{Conclusion}
We proposed a new generative approach that can learn to model transferable target-specific perturbations. Given an image from any source class, our approach can synthesize perturbations that lead to its misclassification on a variety of black-box target models. The core of our approach is an instance-adaptive generator function that is learned using a novel loss formulation. Our loss 
focuses on matching the distribution-level statistics of perturbed source and target samples. By its design, our approach can work with both supervised and unsupervised representations. We demonstrate impressive transferability rates across a range of attack settings compared to state-of-the-art. 
Our results advocate for the use of global loss functions defined over distributions to craft highly transferable adversarial patterns. In future work, we plan to extend proposed method to other model families such as vision transformers \cite{naseer2021improving,naseer2021intriguing,khan2021transformers}.

{\small
\bibliographystyle{ieee_fullname}
\bibliography{egbib}
}
\setcounter{section}{0}
\setcounter{table}{0}
\setcounter{figure}{0}
\def\thesection{Appendix \Alph{section}}
\twocolumn[
  \begin{@twocolumnfalse}
    \begin{center}
        \section*{Supplementary: On Generating Transferable Target Perturbations}
    \end{center}
  \end{@twocolumnfalse}
]

We study the effect of augmentations and ensemble learning by analysing class-wise transferability in ~\ref{sec: sup_per_class_analysis}. We further discuss on why augmentations and ensemble learning leads to more transferable targeted patters in ~\ref{subsec: why_augs} 
\& \ref{subsec: why_ensemble}. We then present the vulnerability of batchnorm to black-box targeted perturbations in  \ref{sec: sup_batchnorm}. In \ref{sec: skip_connections_and_linear_backpropagation}, we analyze the effect of linear back-propagation of gradients \cite{guo2020backpropagating} and using more gradients from skip connections \cite{wu2020skip} on the targeted attack transferability. For the sake of completeness, we report the drop in clean accuracy caused by different defenses including input processing methods (JPEG, Median Blur, and NRP), adversarial training, and stylized training in \ref{sec: sup_clean_acc_vs_defenses}. 
Names of 100 target  classes are provided in \ref{sec: sup_target_names}. Finally, we present visual illustrations to showcase different targeted adversarial patterns found by our method, \ours (Transferable Targeted Perturbations), in \ref{sec: sup_visual_demo}.

\section{Effect of Augmentations and \\ Ensemble Learning }
\label{sec: sup_per_class_analysis}
We proposed a mechanism to explore augmented adversarial space and ensemble learning to boost transferability of the targeted adversarial perturbations found by \textsc{TTP}. A per-class analysis for 10 targets presented in Table \ref{suptab: augs_and_ensl} reveals that augmentations and ensemble learning increase the adversarial effect for every target. \textsc{TTP} is trained against naturally trained ResNet50 and ResNet ensemble \cls{R$_{ens}$:ResNet\{18,50,101,152\}} and perturbations are transferred to naturally trained VGG16 and stylized VGG16 \cite{geirhos2018imagenettrained}. In some cases, such as \cls{Hippopotamus}, augmented learning maximizes the transferability from ResNet50 to naturally trained VGG16 by more than 100\% (Table \ref{suptab: augs_and_ensl}). Similarly, we observe that ensemble learning proves to be effective e.g., see \cls{Grey-Owl} in Table \ref{suptab: augs_and_ensl}. VGG16 trained on stylized ImageNet showed higher resistance against targeted adversarial attacks. For example, transferability of perturbations found by \textsc{TTP} for \cls{French Bulldog} distribution is around 11\% on VGG16 (SIN) as compared to 63\% on VGG16 trained on ImageNet (IN) (Table \ref{suptab: augs_and_ensl}).

\subsection{Why Augmentations boost Transferability?}
\label{subsec: why_augs}
Ilyas etal \cite{NEURIPS2019_e2c420d9}  showed that adversarial examples can be explained by features of the attacked class label. In our targeted attack case, we wish to imprint the features of the target class distribution onto the source samples within an allowed distance (\eg $l_\infty \le 16$). However, black-box (unknown) model might apply different set of transformations (from one layer to another) to process such features and reduce the target transferability. Training on adversarial augmented samples allows the generator to capture such targeted features that are robust to transformations that may vary from one model to another. 

\subsection{Why Ensemble of week Models maximize Transferability?}
\label{subsec: why_ensemble}
\begin{figure}[t]
\centering
  \begin{minipage}{.150\textwidth}
  	\centering
  	 Clean Image
    \includegraphics[ width=\linewidth]{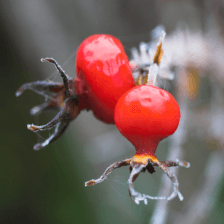}\
    Rosehip
  \end{minipage}
    \begin{minipage}{.150\textwidth}
  	\centering
  Dense121
    \includegraphics[ width=\linewidth]{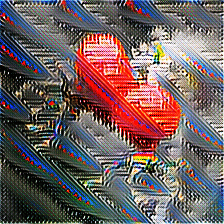}\
    Snowmobile
  \end{minipage}
   \begin{minipage}{.150\textwidth}
  	\centering
  	 Dense169
    \includegraphics[ width=\linewidth]{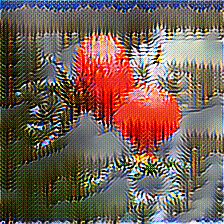}\
    Snowmobile
  \end{minipage}
  \caption{Unconstrained targeted patterns for Snowmobile are shown to demonstrate how discriminators (models) from the same family can capture different information to classify a certain class. Thus, \textsc{TTP} when trained against ensemble of same family models show higher transferability than any of the individual model.}
\label{fig:dense121_vs_dense169_snowmobile}
\end{figure}

Different models of the same family of networks can exploit different information to make prediction. One such example is shown in Fig \ref{fig:dense121_vs_dense169_snowmobile}. Generators are trained against Dense121 and Dense169 to target Snowmobile distribution. Unrestricted generator outputs reveal that Dense121 is more focused on Snowmobile's blades while Dense169 emphasis background pine tree patterns to discriminate Snowmobile samples. This complementary information from different models of the same family helps the generator to capture more generic global patterns for a given target distribution. 

\begin{table*}[h]
	\centering\small
		\setlength{\tabcolsep}{5pt}
		\scalebox{0.95}[0.95]{
		\begin{tabular}{  l|c|cccccccccc|c}
			\toprule
			\rowcolor{Gray}
			\multirow{1}{*}{Source} &\multirow{1}{*}{Augmentations}& \multicolumn{11}{c}{Target Model: VGG16} \\
			\cline{3-13}
			&  & \rotatebox[origin=c]{90}{\cls Grey-Owl} &\rotatebox[origin=c]{90}{\cls Goose}& \rotatebox[origin=c]{90}{\cls Bulldog}&\rotatebox[origin=c]{90}{\cls Hippopotamus}&\rotatebox[origin=c]{90}{\cls Cannon}& \rotatebox[origin=c]{90}{ \cls Fire-Truck}&\rotatebox[origin=c]{90}{\cls Model-T}&\rotatebox[origin=c]{90}{\cls Parachute}&\rotatebox[origin=c]{90}{\cls Snowmobile}&\rotatebox[origin=c]{90}{\cls Street-Sign}&Average\\
			\midrule
			ResNet50 & \xmark&56.5&80.9&49.0&43.9&61.9&82.9&56.5&89.4&41.3&72.9&63.5\\
			ResNet50 & \cmark&56.7&84.1&63.7&94.9&79.5&91.5&76.5&89.8&70.4&80.8&78.8\\
			\cline{3-13}
		    {R$_{ens}$}& \cmark&85.1&94.5&63.3&97.8&90.5&95.8&90.7&96.1&89.6&90.4&89.1\\
		   \midrule
		    \multicolumn{2}{c}{}& \multicolumn{11}{c}{Target Model: VGG16 (SIN)} \\
		    \midrule
			ResNet50 & \xmark&1.61&43.1&0.50&40.9&14.9&9.6&5.8&36.2&6.2&19.2&17.8\\
			ResNet50 & \cmark&1.30&69.6&11.6&68.7&17.0&15.2&20.5&33.2&35.4&30.9&30.3\\
			\cline{3-13}
		   {R$_{ens}$}& \cmark&17.6&77.7&11.4&77.0&59.7&48.4&56.1&72.8&74.1&41.2&53.6\\
			\bottomrule
	\end{tabular}}
    \caption{\emph{\textbf{Per Target Transferability of our Method (\textsc{TTP}):}} Top-1 target accuracy (\%) with 49.95K ImageNet val. samples for each target. Perturbation budget: $l_\infty \le 16$. Adversarial perturbations are transferred from naturally trained ResNet50 and ResNet ensemble to naturally trained VGG16 and stylized VGG16 \cite{geirhos2018imagenettrained}. Augmentations as well as ensemble learning improves efficiency of \textsc{TTP}.}
	\label{suptab: augs_and_ensl}
\end{table*}

\section{The Vulnerability of Batchnorm}
\label{sec: sup_batchnorm}

\begin{figure*}[t]
\centering
  \begin{minipage}{.45\textwidth}
  	\centering
    \includegraphics[ width=\linewidth, keepaspectratio]{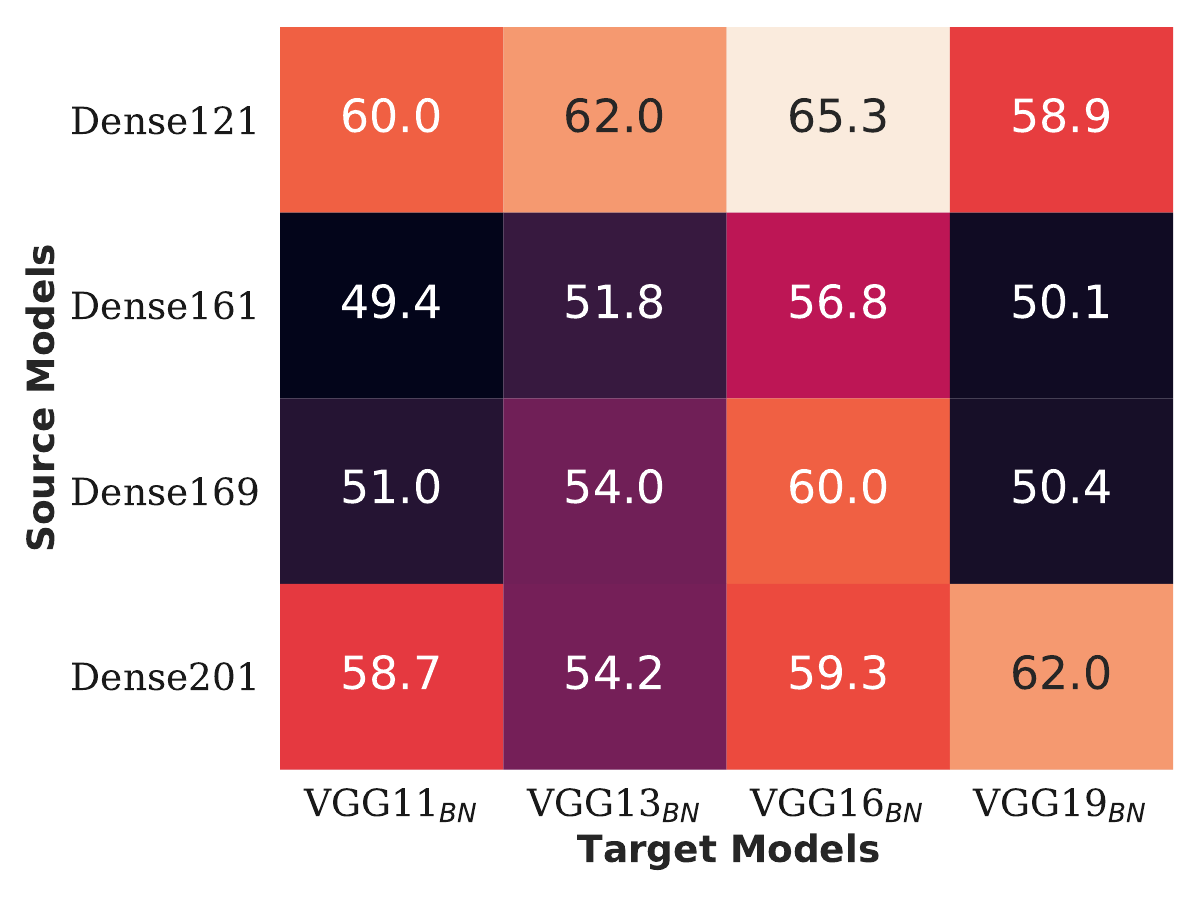}\
  \end{minipage}
    \begin{minipage}{.45\textwidth}
  	\centering
    \includegraphics[width=\linewidth, keepaspectratio]{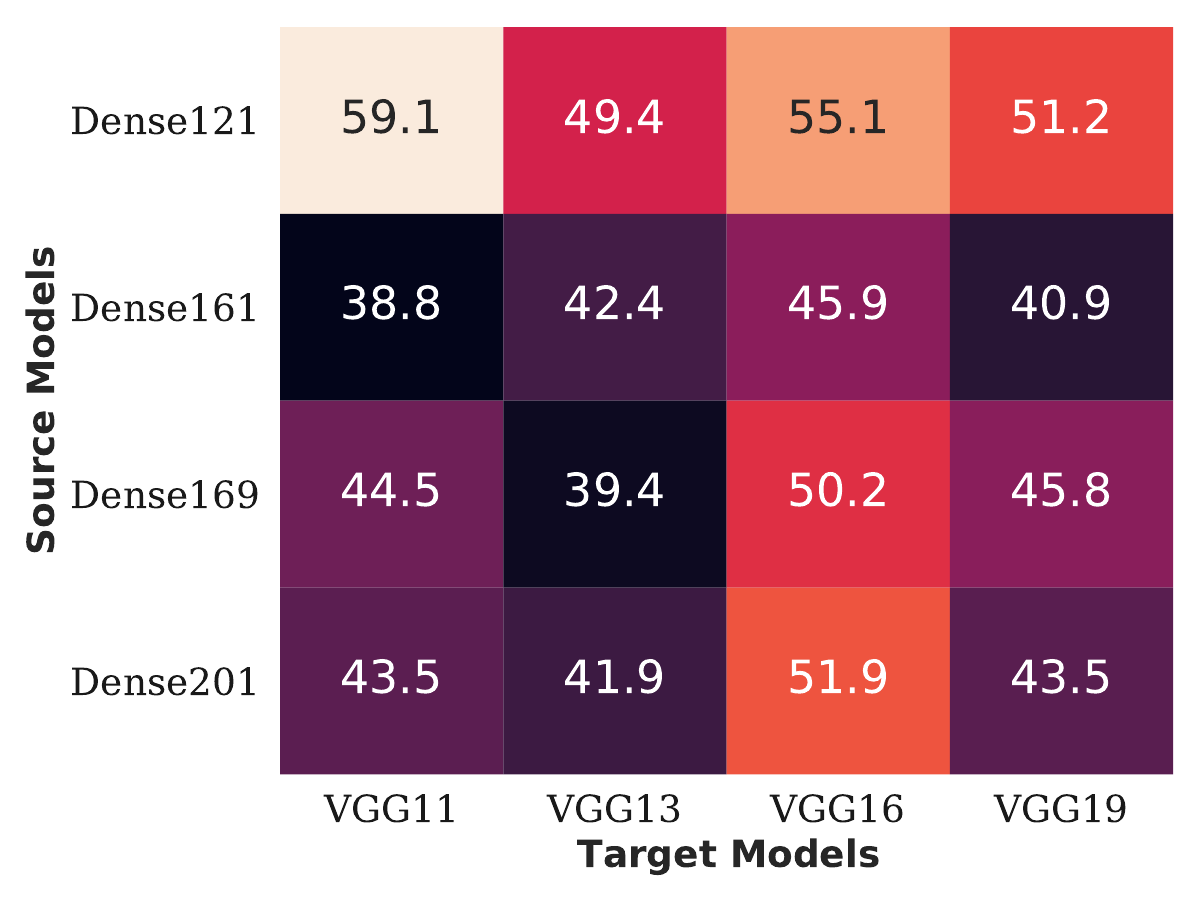}\
  \end{minipage}
  \\
  \begin{minipage}{.45\textwidth}
  	\centering
    \includegraphics[ width=\linewidth, keepaspectratio]{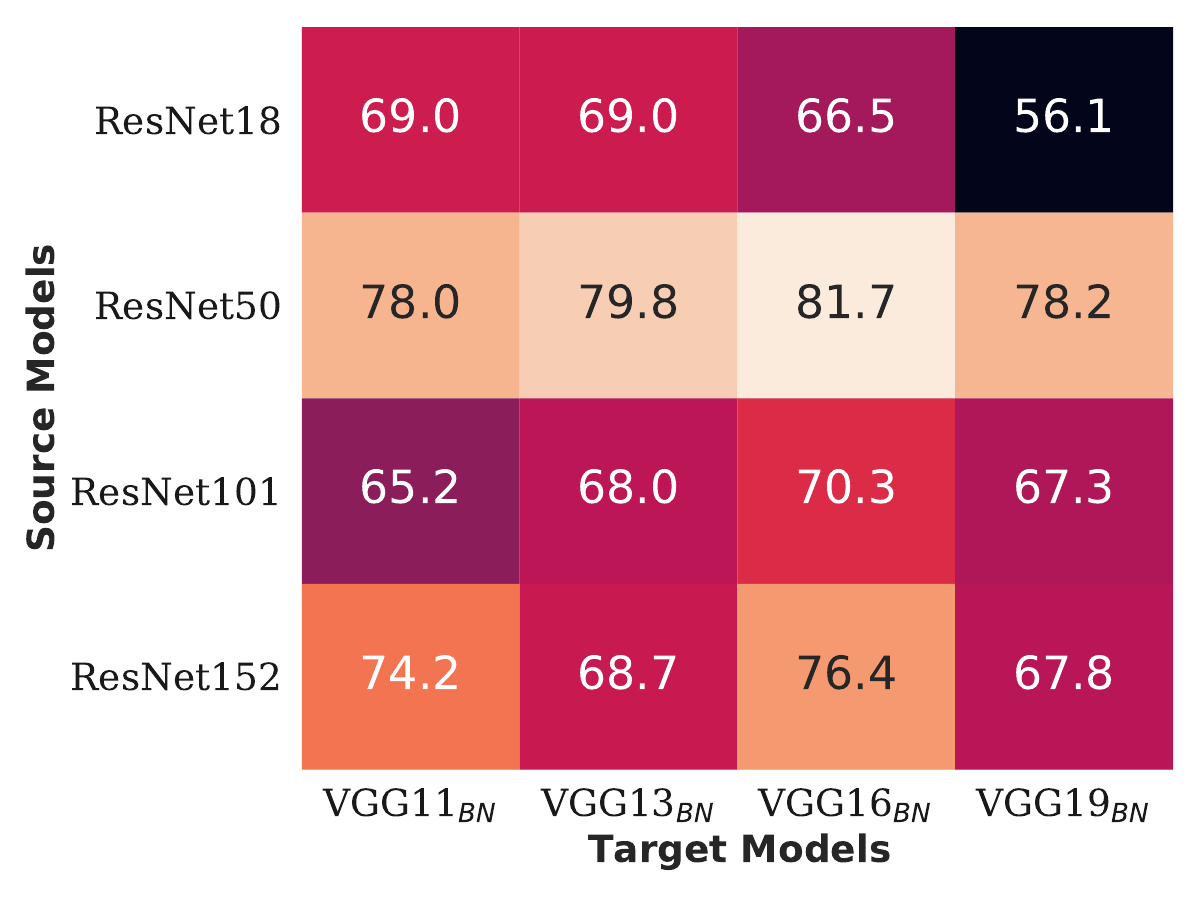}\
  \end{minipage}
    \begin{minipage}{.45\textwidth}
  	\centering
    \includegraphics[width=\linewidth, keepaspectratio]{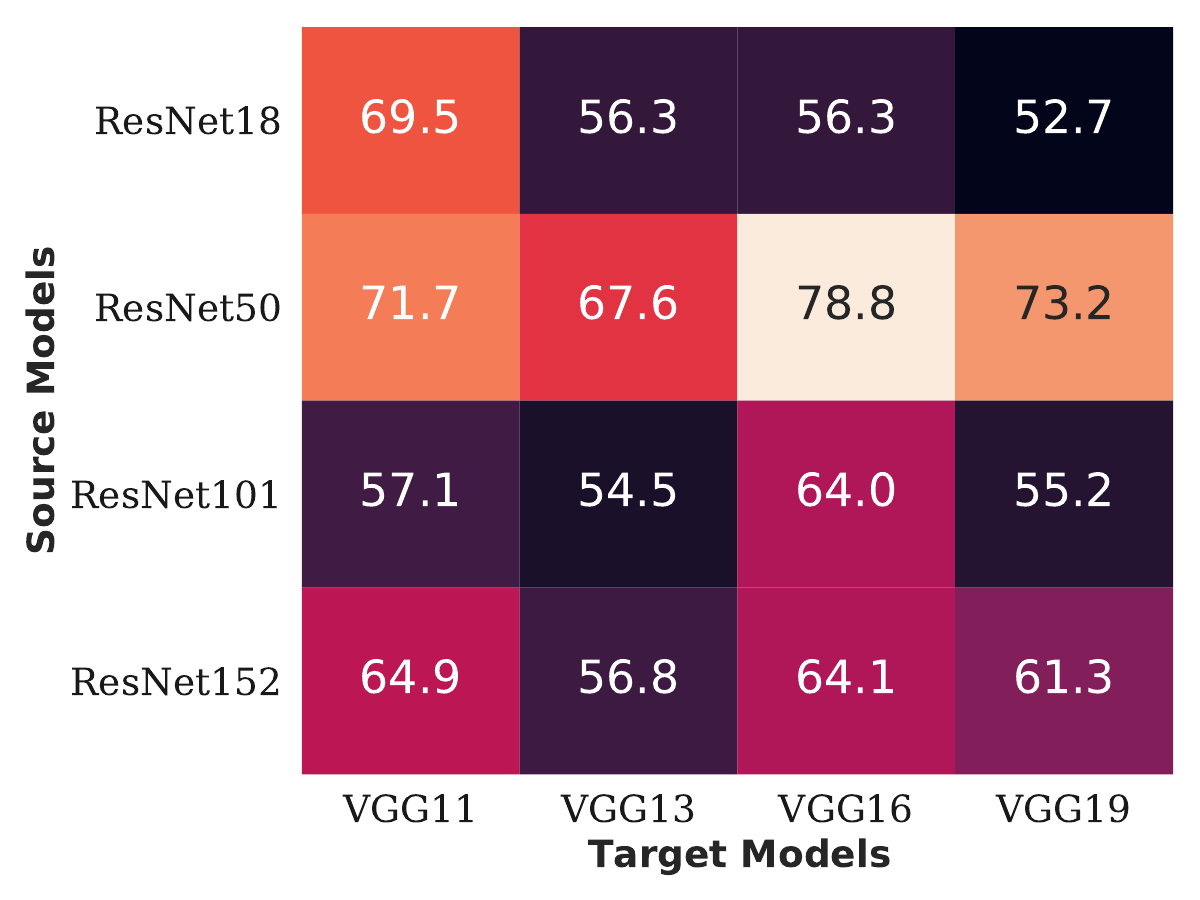}\
  \end{minipage}
\caption{\emph{\textbf{Batchnorm Vulnerability to Targeted Transferability  :}} $\{$10-Targets (all source) settings$\}$. \textsc{TTP} (Algorithm 1 in the paper) strength is higher against models trained naturally with batchnorm as compared to without batchnorm. Batchnorm \cite{ioffe2015batch} provides better optimization and increase model clean accuracy but these empirical results indicate that it also make the model more vulnerable to blackbox targeted attacks. Each value is averaged across 10 targets (see Section 4 in the paper for details) with 49.95k ImageNet val. samples for each target. Perturbation budget is $l_\infty =16$.}
\label{fig: sup_heatmps}

\end{figure*}

Batchnorm \cite{ioffe2015batch} helps in optimization of neural networks as well as increases their clean accuracy. However, our empirical cross-family (Dense $\rightarrow$ VGG$_{BN}$, Dense $\rightarrow$ VGG, ResNet $\rightarrow$ VGG$_{BN}$, ResNet $\rightarrow$ VGG) analysis presented in Fig.~\ref{fig: sup_heatmps} suggests that batchnorm makes the model more vulnerable to the targeted adversarial attacks. Adversarial perturbations found by \textsc{TTP} transfer better against models trained using batchnorm as compared to models trained without it (Fig.~\ref{fig: sup_heatmps}).

\begin{table*}[!t]
	\centering \small
		\setlength{\tabcolsep}{7pt}
		\scalebox{0.8}[0.8]{
		\begin{tabular}{ l|l|ccccc|c|cc|cc}
			\toprule
			\rowcolor{Gray}
			Source& Attack& \multicolumn{5}{c}{Natural Training} & \multicolumn{1}{c}{Augs.} & \multicolumn{2}{c}{Stylized} & \multicolumn{2}{c}{Adversarial} \\
			\cline{3-12}
			& & \rotatebox[origin=c]{90}{{VGG19$_{BN}$}} & \rotatebox[origin=c]{90}{Dense121} &  \rotatebox[origin=c]{90}{ResNet152} & \rotatebox[origin=c]{90}{WRN-50-2} & \rotatebox[origin=c]{90}{VGG16}& \rotatebox[origin=c]{90}{Augmix} & \rotatebox[origin=c]{90}{SIN}  &  \rotatebox[origin=c]{90}{VGG16 (SIN)} & \rotatebox[origin=c]{90}{Adv. ($l_\infty=0.5$)} & \rotatebox[origin=c]{90}{Adv. ($l_\infty=1.0$)}\\
			\midrule
			\multirow{6}{*}{\rotatebox[origin=c]{90}{ResNet50}} & \\
			&PGD \cite{madry2018towards} & 0.8/2.1&1.9/3.7&3.0/4.7&2.5/4.4&0.3/1.5&0.4/1.3&0.1/0.4&0.1/0.3&0.0/0.0&0.0/0.0\\
			&MI \cite{dong2018boosting}&1.5/1.8&3.2/6.2&3.1/5.6&3.0/4.6&1.1/1.4&1.0/1.6&0.3/0.9&0.2/0.4&0.0/0.1&0.0/0.0\\
			&DIM \cite{xie2019improving} &10.4/14.4& 16.2/26.0&13.4/20.9&13.4/19.8&6.4/6.7&4.8/7.7&1.7/3.2&0.5/1.2&0.2/0.5&0.1/0.1\\
			&Po-TRIP \cite{li2020towards}&12.5/15.0& 18.2/30.0&15.9/23.7&14.2/22.3&7.3/8.9&5.5/9.0&2.1/3.7&0.8/2.0&0.3/0.7&0.1/0.1\\
			& FDA-fd \cite{Inkawhich2020Transferable}&16.0/25.3&21.0/33.1&19.7/32.9&17.1/28.4&12.0/18.7 &15.3/19.3&3.1/6.3&1.2/3.0&0.1/1.9&0.1/0.3\\
			& FDA-N \cite{inkawhich2020perturbing}&32.1/38.6&48.3/52.3&37.5/39.0&35.5/40.7&19.0/28.3 &20.3/30.3&5.0/16.6&3.0/10.7&0.6/4.7&0.2/0.8\\
			& SGM \cite{wu2020skip} & 19.2/26.3 &25.9/40.6&19.7/31.1&21.6/30.4&13.5/13.7 & 10.5/15.9 & 2.6/6.1 & 1.3/2.8 & 0.5/1.2 & 0.1/0.3\\
			& SGM \cite{wu2020skip} + LinBP \cite{guo2020backpropagating} & 22.0/27.1& 34.5/40.0& 30.5/32.9&25.1/21.0&14.8/15.0&17.3/25.3&4.6/14.3&2.4/8.0&0.3/2.9&0.1/0.3\\
			\cline{2-12}
			& Ours (\textsc{TTP}) &  \textbf{79.0/81.4}&\textbf{84.4/87.0}&\textbf{81.9/86.6}&\textbf{80.2/81.2}&\textbf{79.4/78.2}&\textbf{72.7/81.2}&\textbf{30.5/42.4}&\textbf{29.3/36.9}&\textbf{5.5/50.1}&\textbf{0.4/17.1}\\
			\bottomrule
	\end{tabular}}
	\caption{\emph{\textbf{Target Transferability:}} $\{$10-Targets (sub-source)$\}$ Top-1 target accuracy (\%) averaged across 10 targets. Perturbation budget: $l_\infty \le 16/32$. SIN \cite{geirhos2018imagenettrained} and Adv ($l_\infty$=0.5), and  Adv ($l_\infty$=1.0) \cite{salman2020adversarially} are ResNet50 models trained using stylized and adversarial examples, respectively. Augs. represents augmentation based training \cite{hendrycks2020augmix} of ResNet50. }
	\label{tab: suppl_skip_linear}
\end{table*}

\begin{table}[t]
    \centering
        \begin{tabular}{c|c|c|c}
    \toprule
    \rowcolor{Gray}
         Model & Defense & Accuracy & Difference \\
         \midrule
         \multirow{4}{*}{VGG19$_{BN}$}    &  -- &74.24& 0.0\\
         & JPEG&67.34&-6.90 \\
         & Blur & 53.86& -20.38\\
         & NRP & 72.00 & -2.24\\
         \midrule
        \multirow{4}{*}{Dense121}    &  -- & 74.65 & 0.0\\
         & JPEG&68.92&-5.73 \\
         & Blur & 61.27&-13.38 \\
         & NRP & 72.01 & -2.63 \\
         \midrule
        \multirow{4}{*}{ResNet50}    &  -- & 76.15 & 0.0\\
         & JPEG &70.82&-5.33\\
         & Blur & 61.30&-14.85\\
         & NRP & 73.21 & -2.94 \\
    \bottomrule
    \end{tabular}
    \caption{\emph{\textbf{Effect of Input Processing on Clean Accuracy:}} Top-1 (\%) accuracy on ImageNet val. set (50k images). Median Blur with window size 5$\times$5 causes large drop in clean accuracy while NRP \cite{Naseer_2020_CVPR} has the least effect on the model's clean accuracy.}
    \label{tab: drop_in_acc_with_input_processing}
\end{table}

\begin{table}[t]
    \centering
    \begin{tabular}{c|c|c|c}
    \toprule
    \rowcolor{Gray}
         Model & Training Type & Accuracy & Difference \\
         \midrule
         \multirow{8}{*}{ResNet50}    &  IN & 76.15 & 0.0\\
         & SIN & 60.18 & -15.97\\
         & SIN-IN&74.59&-1.56 \\
         & Augmix& 77.53& +1.38\\
         & Adv. ($l_\infty$, $\epsilon=.5$)&73.73&-2.42\\
         & Adv. ($l_\infty$, $\epsilon=1$)&72.05&-4.10\\
         & Adv. ($l_2$, $\epsilon=.1$)&74.78&-1.37\\
         & Adv. ($l_2$, $\epsilon=.5$)&73.16&-2.99\\
         \midrule
        \multirow{2}{*}{VGG16}    &  IN & 71.59 & 0.0 \\
         & SIN & 52.26 & -19.33\\
    \bottomrule
    \end{tabular}
    \caption{\emph{\textbf{Effect of Robust Training on Clean Accuracy:}} Top-1 (\%) accuracy on ImageNet val. set (50k images). Every training mechanism with the exception of Augmix \cite{hendrycks2020augmix} reduces model's clean accuracy. Stylized training \cite{geirhos2018imagenettrained} causes significant drop in accuracy in comparison to other types of training methods.}
    \label{tab: drop_in_acc_with_different_training}
\end{table}

\section{Skip Connections and Linear Back-Propagation of Gradients}
\label{sec: skip_connections_and_linear_backpropagation}
Dongxian \etal \cite{wu2020skip} observed that while back-propagating, giving more importance to the gradients coming from skip connections can enhance adversarial transferabililty. Similarly, Guo \etal \cite{guo2020backpropagating} showed that encouraging linearity while back-propagating gradients improve transferability. Here, we analyze target transferability of both of these techniques \cite{wu2020skip, guo2020backpropagating} and present a holistic comparison between all the considered iterative and generative attacks in Table \ref{tab: suppl_skip_linear}. Our approach sets new state-of-the-art.

\section{Clean Accuracy vs. Defenses}
\label{sec: sup_clean_acc_vs_defenses}
We evaluate the effect of different defenses on model's clean accuracy. We study the input processing methods including JPEG with quality 50\% \cite{naseer2019local}, Median Blur with kernel size 5$\times$5 \cite{naseer2019local} and NRP \cite{Naseer_2020_CVPR} as well as different training mechanisms including Augmix \cite{hendrycks2020augmix}, stylized \cite{geirhos2018imagenettrained} and adversarial training methods \cite{madry2018towards, salman2020adversarially}. Results are presented in Tables \ref{tab: drop_in_acc_with_input_processing} \& \ref{tab: drop_in_acc_with_different_training}. We observe that Median Blur causes a significant drop in clean accuracy (Table~\ref{tab: drop_in_acc_with_input_processing}) while among training methods, stylized training (SIN) \cite{geirhos2018imagenettrained} has the most negative effect on the clean accuracy.

\section{100 Targets Names}
\label{sec: sup_target_names}
The performance of \textsc{TTP} is evaluated against the following randomly selected 100 targets (see Sec. 4.1 of the paper). We divide ImageNet classes into 100 mutually exclusive sets. Each set contains 10 classes. We randomly selected one target from each set.

\noindent \cls{Tiger-Shark}, \cls{Bulbul}, \cls{Grey-Owl}, \cls{Terrapin}, \cls{Komodo-Dragon}, \cls{Thunder-Snake}, \cls{Trilobite}, \cls{Scorpion}, \cls{Quail}, \cls{Goose}, \cls{Jellyfish}, \cls{Slug}, \cls{Flamingo}, \cls{Bustard}, \cls{Dowitcher}, \cls{Chihuahua}, \cls{Beagle}, \cls{Weimaraner}, \cls{Lakeland-Terrier}, \cls{Australian-Terrier}, \cls{Golden-Retriever}, \cls{English-Setter}, \cls{Komondor}, \cls{Appenzeller}, \cls{French-Bulldog}, \cls{Chow}, \cls{Keeshond}, \cls{Hyaena}, \cls{Egyptian-Cat}, \cls{Lion}, \cls{Bee}, \cls{Leafhopper}, \cls{Sea-Urchin}, \cls{Zebra}, \cls{Hippopotamus}, \cls{Polecat}, \cls{Gorilla}, \cls{Langur}, \cls{Eel}, \cls{Anemone-Fish}, \cls{Airliner}, \cls{Banjo}, \cls{Bassinet}, \cls{Beaker}, \cls{Bell-Cote}, \cls{Bookcase}, \cls{Buckle}, \cls{Cannon}, \cls{CD-Player}, \cls{Chain-Saw}, \cls{Coil}, \cls{Cornet}, \cls{Crutch}, \cls{Dome}, \cls{Electric-Guitar}, \cls{Fire-Truck}, \cls{Garbage-Truck},\cls{ Greenhouse}, \cls{Grocery-Store}, \cls{Honeycomb}, \cls{iPod}, \cls{Jigsaw-Puzzle}, \cls{Lipstick}, \cls{Maillot}, \cls{Maze}, \cls{Military-Uniform}, \cls{Model-T}, \cls{Neck-Brace}, \cls{Overskirt}, \cls{Parachute}, \cls{Pay-Phone}, \cls{Pickup}, \cls{Pirate-Ship}, \cls{Poncho}, \cls{Purse}, \cls{Rain-Barrel}, \cls{Rotisserie}, \cls{School-Bus}, \cls{Sewing-Machine}, \cls{Shopping-Cart}, \cls{Snowmobile}, \cls{Spatula}, \cls{Stove}, \cls{Sunglass}, \cls{Teapot}, \cls{Toaster}, \cls{Tractor}, \cls{Umbrella}, \cls{Velvet}, \cls{Wallet}, \cls{Whiskey-Jug}, \cls{Street-Sign}, \cls{Ice-Lolly}, \cls{Pretzel}, \cls{Cardoon}, \cls{Hay}, \cls{Pizza}, \cls{Volcano}, \cls{Rapeseed}, \cls{Agaric}

\section{Visual Demos}
\label{sec: sup_visual_demo}
\begin{figure*}[t]
\centering
\includegraphics[width=\linewidth, keepaspectratio]{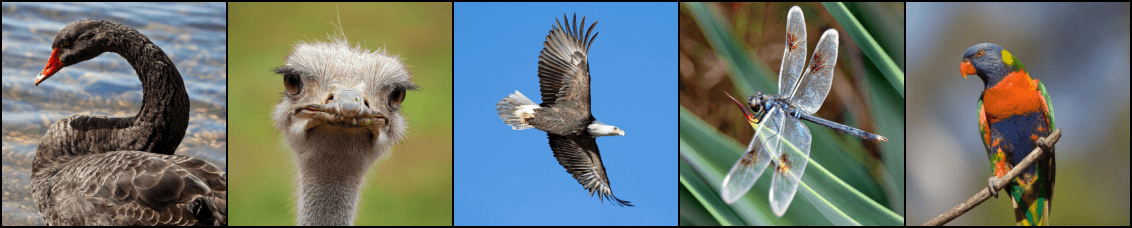}
Original Images
\includegraphics[width=\linewidth, keepaspectratio]{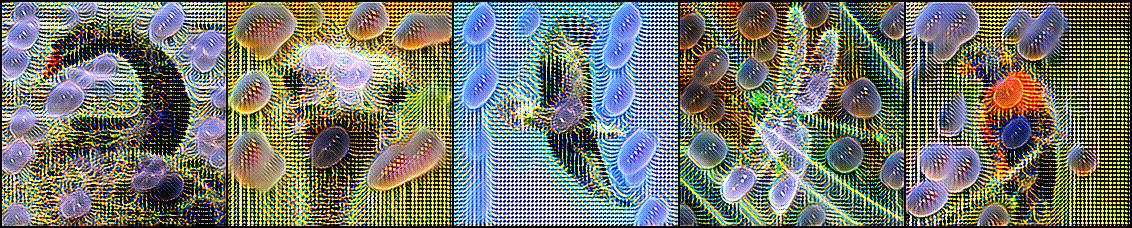}
\includegraphics[width=\linewidth, keepaspectratio]{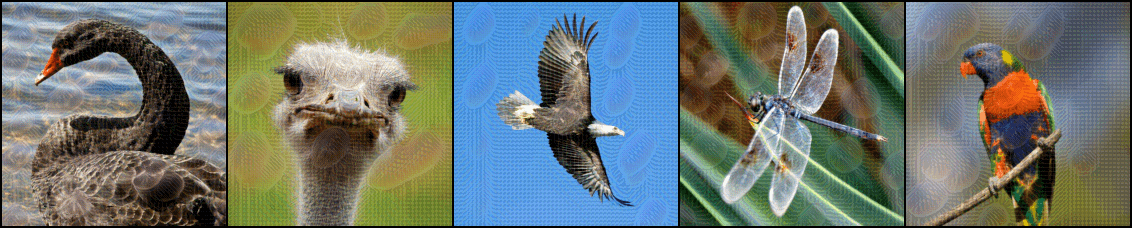}
Source model: ResNet50, Target Distribution: Jellyfish, Transferabiliy to Dense121: 90.05 \%
\includegraphics[width=\linewidth, keepaspectratio]{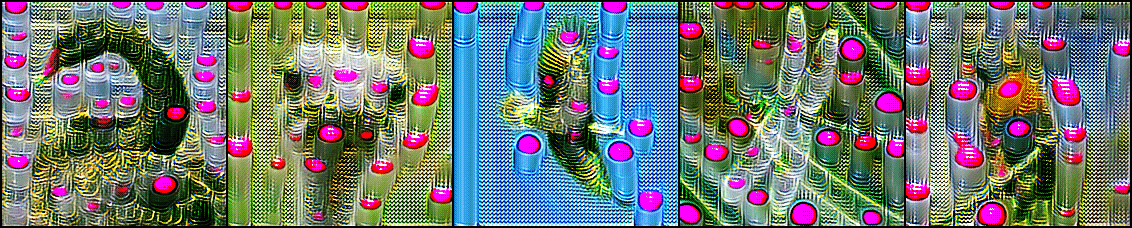}
\includegraphics[width=\linewidth, keepaspectratio]{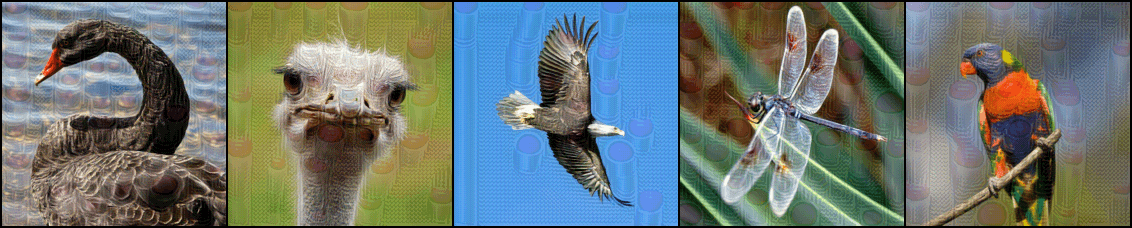}
Source model: ResNet50, Target Distribution: Lipstick, Transferability to Dense121: 95.20  \%
 
\caption{Targeted adversaries produced by \ours (before and after valid projection) trained against ResNet50. Observe that adversarial patterns are not constant rather \ours adapts to the input sample and adds different patterns to different samples to achieve maximum transferability. Transferability is measured as Top-1 target accuracy on the ImageNet val. set (49.95k samples excluding the target images).}
 \label{fig:1st_page_main_demo}
\end{figure*}

\begin{figure*}[t]
\centering
\includegraphics[width=\linewidth, keepaspectratio]{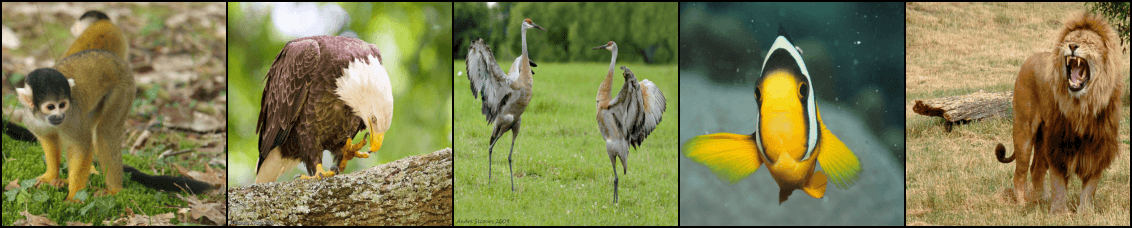}
Original Images
\includegraphics[width=\linewidth, keepaspectratio]{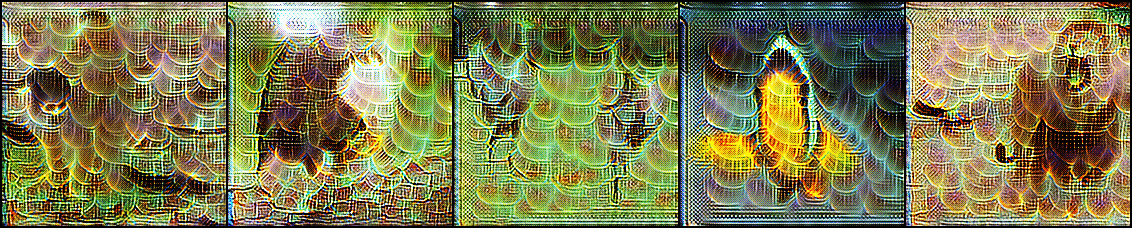}
\includegraphics[width=\linewidth, keepaspectratio]{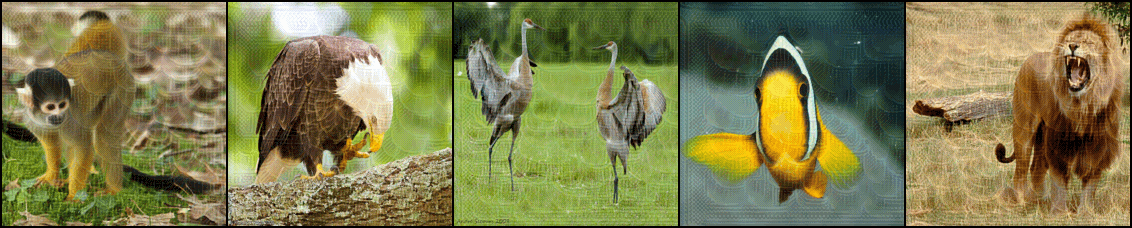}
Source model: ResNet50, Target Distribution: Stove, Transferabiliy to Dense121: 36.86\% 
\includegraphics[width=\linewidth, keepaspectratio]{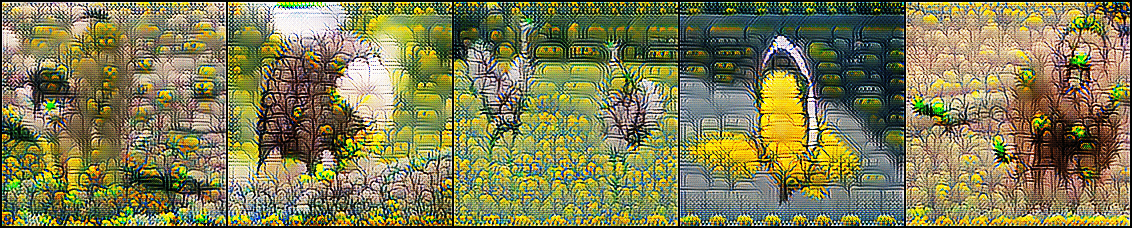}
\includegraphics[width=\linewidth, keepaspectratio]{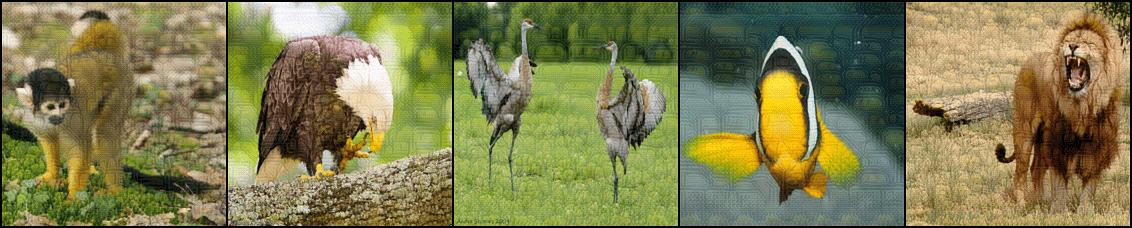}
Source model: ResNet50, Target Distribution: Rapeseed, Transferabiliy to Dense121: 49.59\% 
 
\caption{Targeted adversaries produced by \ours (before and after valid projection) trained against ResNet50. Observe that adversarial patterns are not constant rather \ours adapts to the input sample and adds different patterns to different samples to achieve maximum transferability. Transferability is measured as Top-1 target accuracy on the ImageNet val. set (49.95k samples excluding the target images).}
 \label{fig:2st_page_main_demo}
\end{figure*}

\begin{figure*}[ht]
\centering
\includegraphics[width=\linewidth, keepaspectratio]{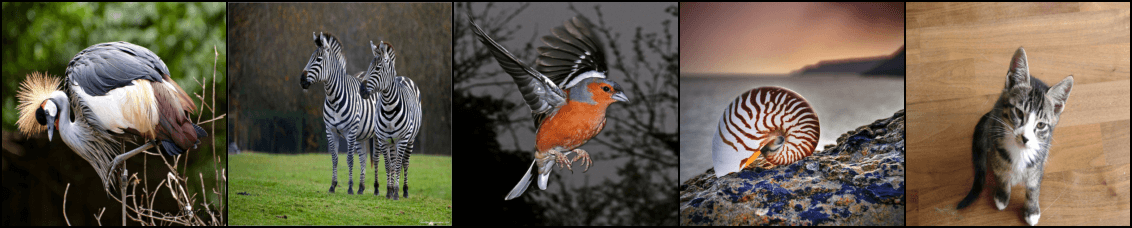}
Original Images
\includegraphics[width=\linewidth, keepaspectratio]{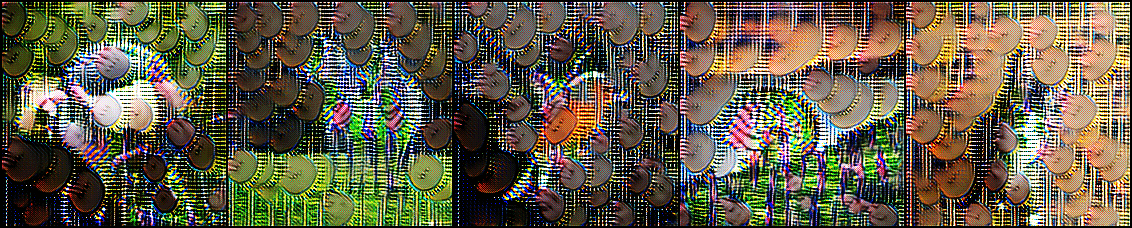}
\includegraphics[width=\linewidth, keepaspectratio]{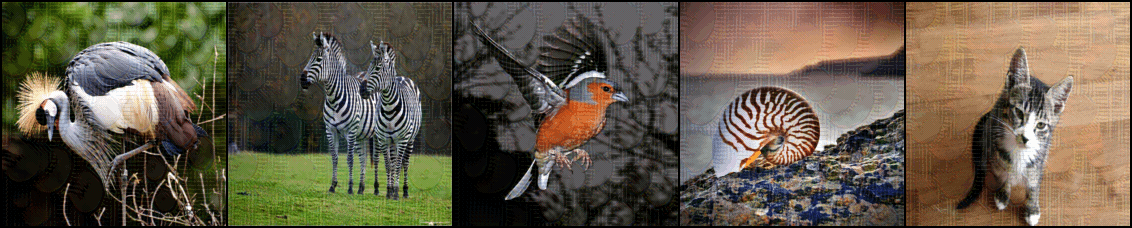}
Source model: ResNet50, Target Distribution: Banjo, Transferabiliy to Dense121: 82.95\% 
\includegraphics[width=\linewidth, keepaspectratio]{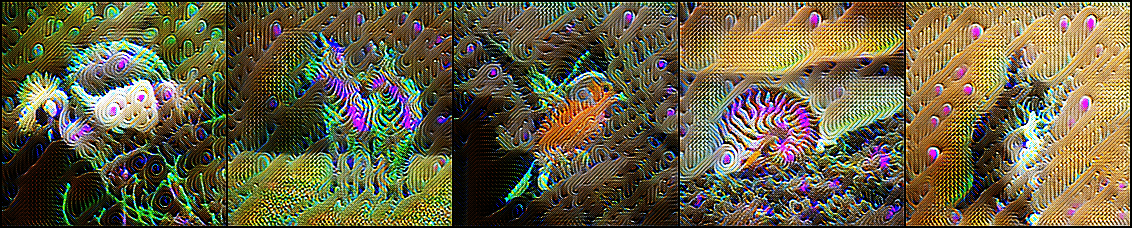}
\includegraphics[width=\linewidth, keepaspectratio]{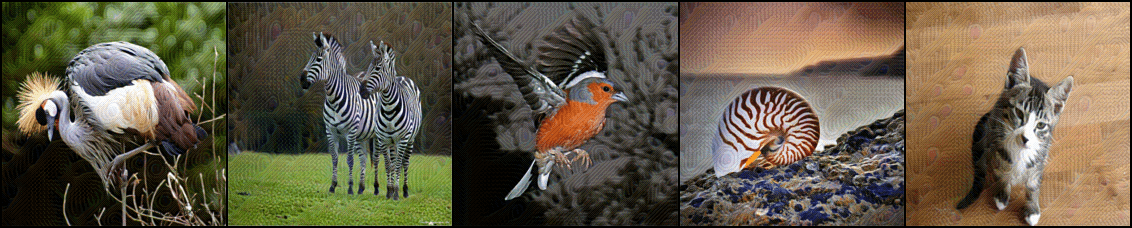}
Source model: ResNet50, Target Distribution: Anemone Fish, Transferabiliy to Dense121: 74.45\% 
 
\caption{Targeted adversaries produced by \ours (before and after valid projection) trained against ResNet50. Observe that adversarial patterns are not constant rather \ours adapts to the input sample and adds different patterns to different samples to achieve maximum transferability. Transferability is measured as Top-1 target accuracy on the ImageNet val. set (49.95k samples excluding the target images).}
 \label{fig:3st_page_main_demo}
\end{figure*}

\begin{figure*}[ht]
\centering
\includegraphics[width=\linewidth, keepaspectratio]{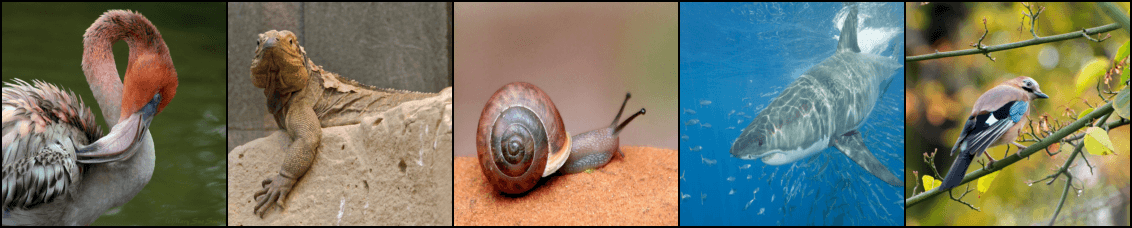}
Original Images
\includegraphics[width=\linewidth, keepaspectratio]{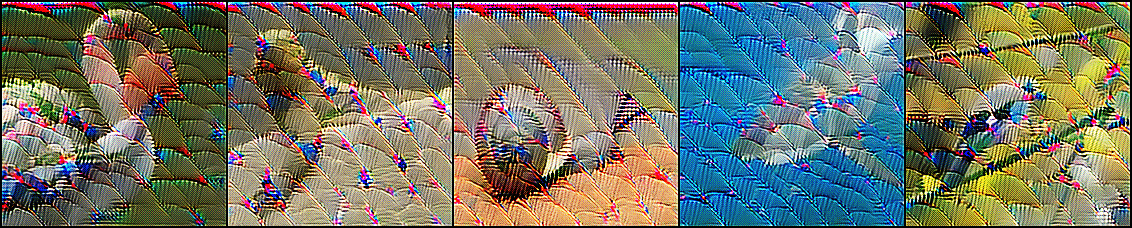}
\includegraphics[width=\linewidth, keepaspectratio]{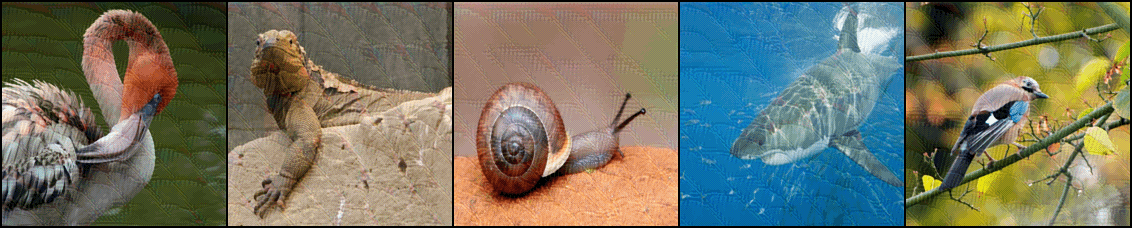}
Source model: ResNet50, Target Distribution: Parachute, Transferabiliy to Dense121: 95.30\% 
\includegraphics[width=\linewidth, keepaspectratio]{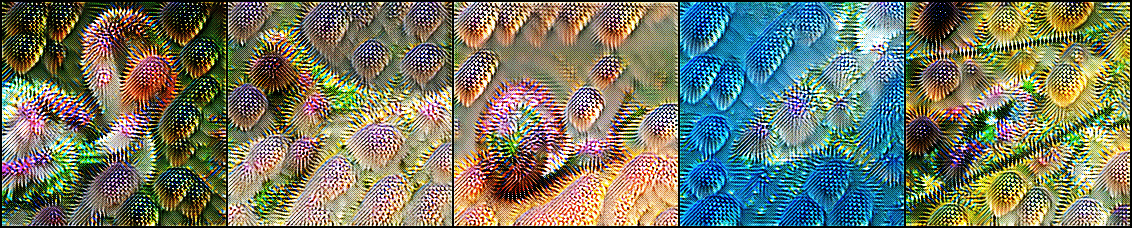}
\includegraphics[width=\linewidth, keepaspectratio]{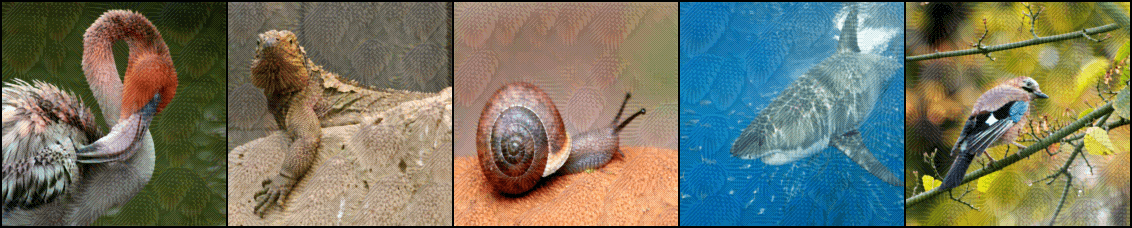}
Source model: ResNet50, Target Distribution: Sea Urchin, Transferabiliy to Dense121: 89.10\% 
 
\caption{Targeted adversaries produced by \ours (before and after valid projection) trained against ResNet50. Observe that adversarial patterns are not constant rather \ours adapts to the input sample and adds different patterns to different samples to achieve maximum transferability. Transferability is measured as Top-1 target accuracy on the ImageNet val. set (49.95k samples excluding the target images).}
 \label{fig:4st_page_main_demo}
\end{figure*}

\begin{figure*}[ht]
\centering
\includegraphics[width=\linewidth, keepaspectratio]{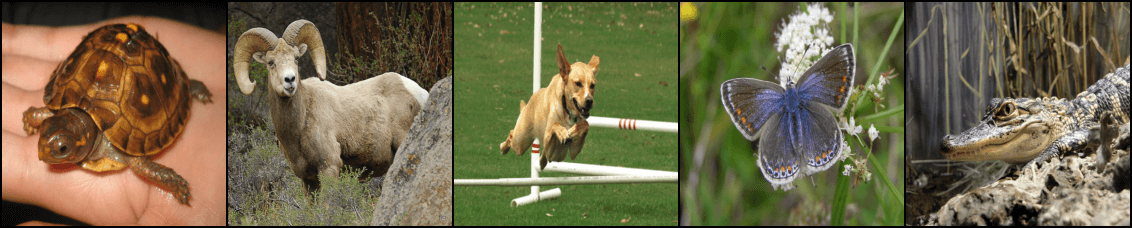}
Original Images
\includegraphics[width=\linewidth, keepaspectratio]{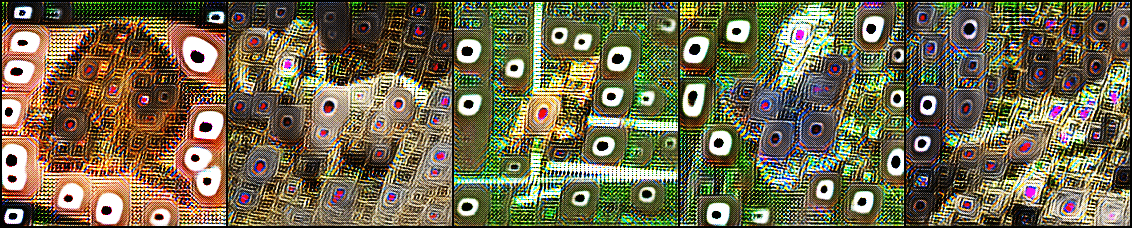}
\includegraphics[width=\linewidth, keepaspectratio]{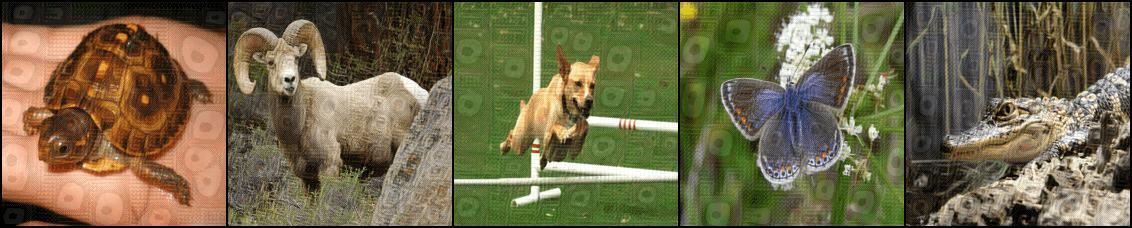}
Source model: ResNet50, Target Distribution: iPOD, Transferabiliy to Dense121: 69.86\% 
\includegraphics[width=\linewidth, keepaspectratio]{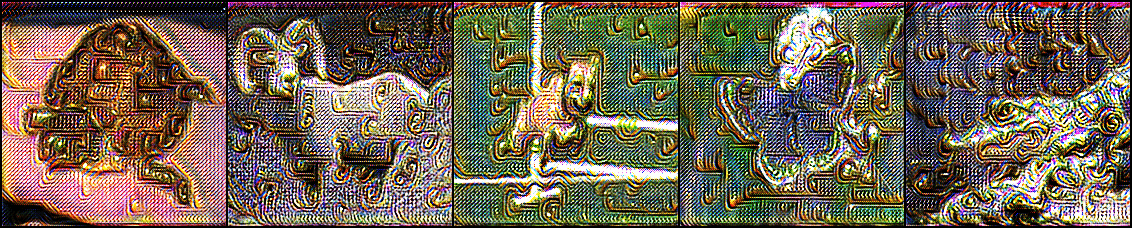}
\includegraphics[width=\linewidth, keepaspectratio]{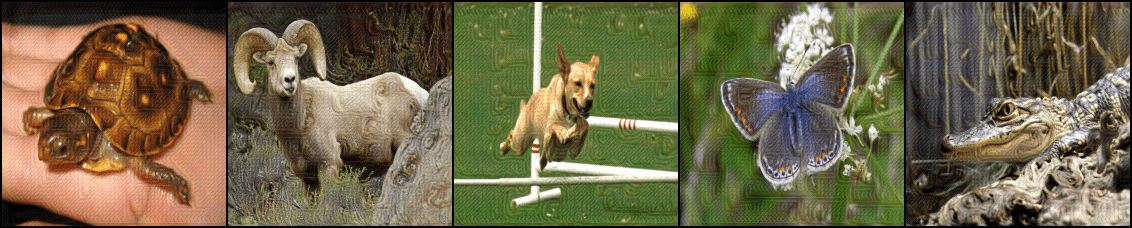}
Source model: ResNet50, Target Distribution: Buckle, Transferabiliy to Dense121: 77.06\% 
 
\caption{Targeted adversaries produced by \ours (before and after valid projection) trained against ResNet50. Observe that adversarial patterns are not constant rather \ours adapts to the input sample and adds different patterns to different samples to achieve maximum transferability. Transferability is measured as Top-1 target accuracy on the ImageNet val. set (49.95k samples excluding the target images).}
 \label{fig:5st_page_main_demo}
\end{figure*}

\begin{figure*}[ht]
\centering
\includegraphics[width=\linewidth, keepaspectratio]{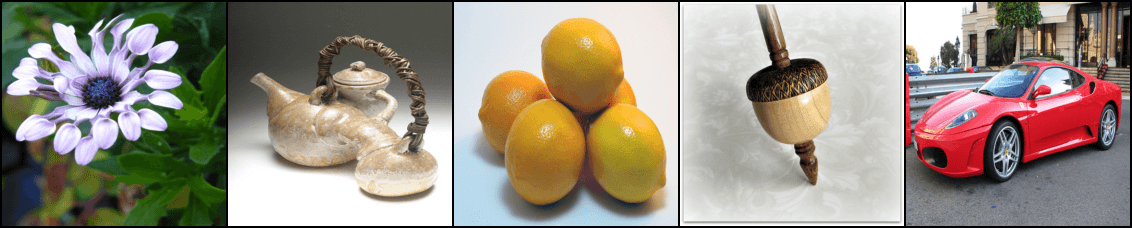}
Original Images
\includegraphics[width=\linewidth, keepaspectratio]{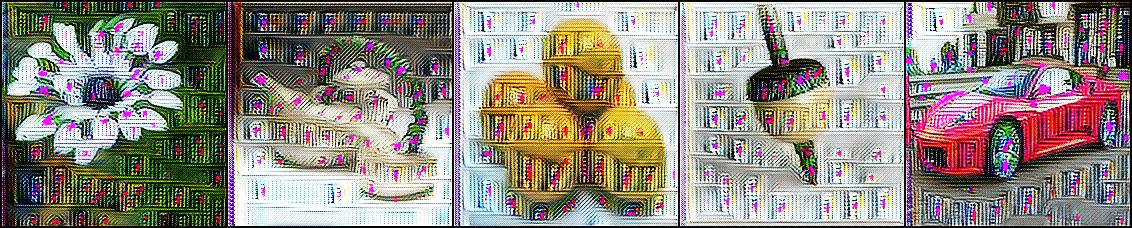}
\includegraphics[width=\linewidth, keepaspectratio]{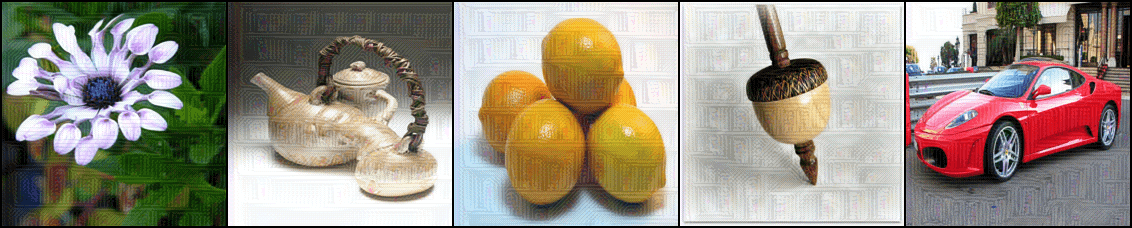}
Source model: ResNet50, Target Distribution: Bookcase, Transferabiliy to Dense121: 85.21\% 
\includegraphics[width=\linewidth, keepaspectratio]{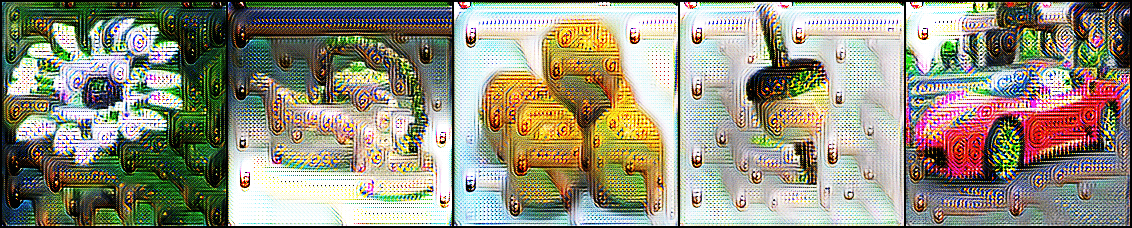}
\includegraphics[width=\linewidth, keepaspectratio]{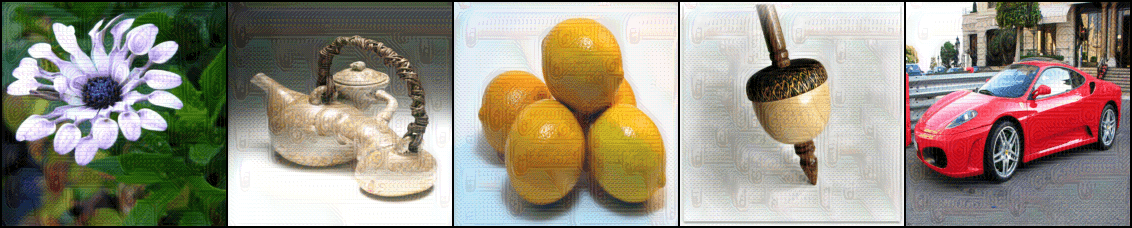}
Source model: ResNet50, Target Distribution: Sewing Machine, Transferabiliy to Dense121: 67.26\% 
 
\caption{Targeted adversaries produced by \ours (before and after valid projection) trained against ResNet50. Observe that adversarial patterns are not constant rather \ours adapts to the input sample and adds different patterns to different samples to achieve maximum transferability. Transferability is measured as Top-1 target accuracy on the ImageNet val. set (49.95k samples excluding the target images).}
 \label{fig:6st_page_main_demo}
\end{figure*}

\begin{figure*}[t]
\centering
  \begin{minipage}{.18\textwidth}
  	\centering
  	 Clean Image
    \includegraphics[ width=\linewidth]{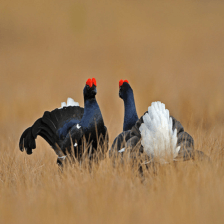}\
    Ptarmigan
  \end{minipage}
    \begin{minipage}{.18\textwidth}
  	\centering
  	 VGG11
    \includegraphics[ width=\linewidth]{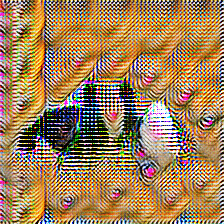}\
    French Bulldog
  \end{minipage}
   \begin{minipage}{.18\textwidth}
  	\centering
  	 VGG13
    \includegraphics[ width=\linewidth]{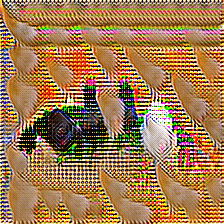}\
    French Bulldog
  \end{minipage}
    \begin{minipage}{.18\textwidth}
  	\centering
  	 VGG16
    \includegraphics[ width=\linewidth]{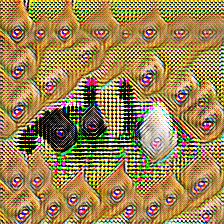}\
    French Bulldog
  \end{minipage}
\begin{minipage}{.18\textwidth}
  	\centering
  	 VGG19
    \includegraphics[ width=\linewidth]{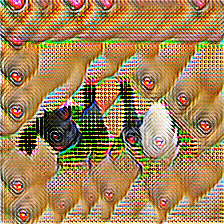}\
    French Bulldog
  \end{minipage}
  \vspace{0.5cm}
  \\
   \begin{minipage}{.18\textwidth}
  	\centering
  	 Clean Image
    \includegraphics[ width=\linewidth]{figures/supplimentary/evolution/original_dense.png}\
    Rosehip
  \end{minipage}
    \begin{minipage}{.18\textwidth}
  	\centering
  	 Dense121
    \includegraphics[ width=\linewidth]{figures/supplimentary/evolution/unrestricted_adv_dense121_802.png}\
    Snowmobile
  \end{minipage}
   \begin{minipage}{.18\textwidth}
  	\centering
  	 Dense161
    \includegraphics[ width=\linewidth]{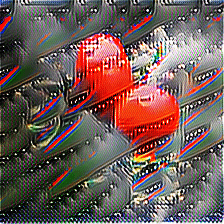}\
    Snowmobile
  \end{minipage}
    \begin{minipage}{.18\textwidth}
  	\centering
  	 Dense169
    \includegraphics[ width=\linewidth]{figures/supplimentary/evolution/unrestricted_adv_dense169_802.png}\
    Snowmobile
  \end{minipage}
\begin{minipage}{.18\textwidth}
  	\centering
  	Dense201
    \includegraphics[ width=\linewidth]{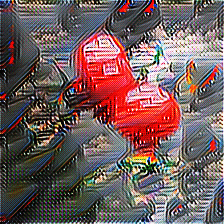}\
    Snowmobile
  \end{minipage}
  \vspace{0.5cm}
  \\ 
 
  \begin{minipage}{.18\textwidth}
  	\centering
  	 Clean Image
    \includegraphics[ width=\linewidth]{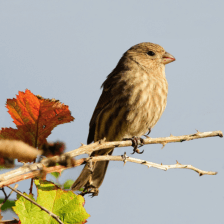}\
    House Finch
  \end{minipage}
    \begin{minipage}{.18\textwidth}
  	\centering
  	 ResNet18
    \includegraphics[ width=\linewidth]{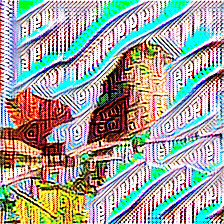}\
     Street Sign
  \end{minipage}
   \begin{minipage}{.18\textwidth}
  	\centering
  	 ResNet50
    \includegraphics[ width=\linewidth]{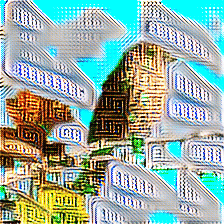}\
     Street Sign
  \end{minipage}
    \begin{minipage}{.18\textwidth}
  	\centering
  	 ResNet101
    \includegraphics[ width=\linewidth]{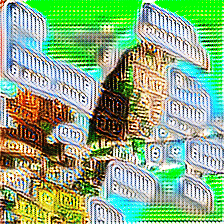}\
     Street Sign
  \end{minipage}
\begin{minipage}{.18\textwidth}
  	\centering
  	 ResNet152
    \includegraphics[ width=\linewidth]{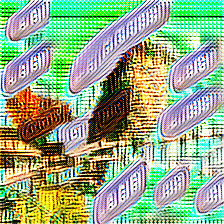}\
    Street Sign
  \end{minipage}
  \caption{\emph{\textbf{Evolution of \textsc{TTP}:}} Unconstrained targeted adversarial patterns generated by \textsc{TTP} are shown to demonstrate how \textsc{TTP} evolves as it learns perturbations from different source models of a certain family of networks.}
\label{fig:sup_evolution_of_patterns_1}
\end{figure*}

\begin{figure*}[t]
\centering
  \begin{minipage}{.18\textwidth}
  	\centering
  	 Clean Image
    \includegraphics[ width=\linewidth]{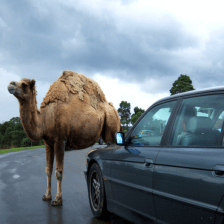}\
    Arabian Camel
  \end{minipage}
    \begin{minipage}{.18\textwidth}
  	\centering
  	 VGG11
    \includegraphics[ width=\linewidth]{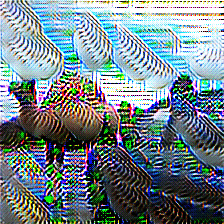}\
    Goose
  \end{minipage}
   \begin{minipage}{.18\textwidth}
  	\centering
  	 VGG13
    \includegraphics[ width=\linewidth]{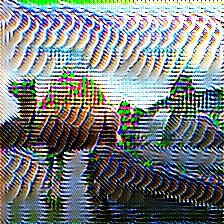}\
    Goose
  \end{minipage}
    \begin{minipage}{.18\textwidth}
  	\centering
  	 VGG16
    \includegraphics[ width=\linewidth]{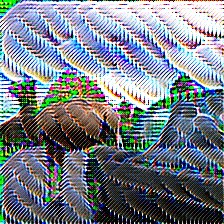}\
    Goose
  \end{minipage}
\begin{minipage}{.18\textwidth}
  	\centering
  	 VGG19
    \includegraphics[ width=\linewidth]{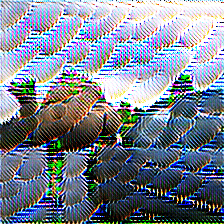}\
    Goose
  \end{minipage}
  \vspace{0.5cm}
  \\
  \begin{minipage}{.18\textwidth}
  	\centering
  	 Clean Image
    \includegraphics[ width=\linewidth]{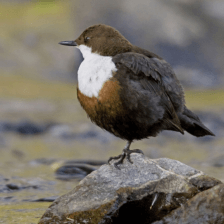}\
    Water Ouzel
  \end{minipage}
    \begin{minipage}{.18\textwidth}
  	\centering
  	 Dense121
    \includegraphics[ width=\linewidth]{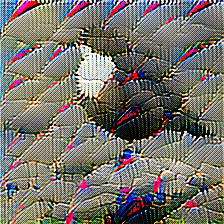}\
    Parachute
  \end{minipage}
  \begin{minipage}{.18\textwidth}
  	\centering
  	 Dense161
    \includegraphics[ width=\linewidth]{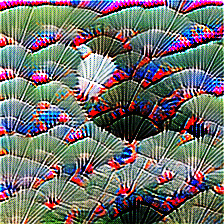}\
    Parachute
  \end{minipage}
    \begin{minipage}{.18\textwidth}
  	\centering
  	 Dense169
    \includegraphics[ width=\linewidth]{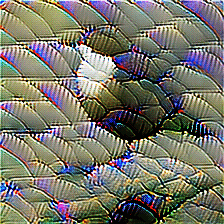}\
    Parachute
  \end{minipage}
\begin{minipage}{.18\textwidth}
  	\centering
  	Dense201
    \includegraphics[ width=\linewidth]{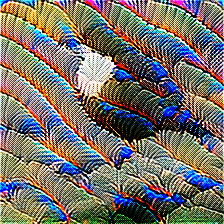}\
    Parachute
  \end{minipage}
  \vspace{0.5cm}
  \\ 
 
  \begin{minipage}{.18\textwidth}
  	\centering
  	 Clean Image
    \includegraphics[ width=\linewidth]{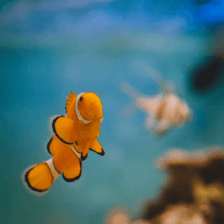}\
   Anemone Fish
  \end{minipage}
    \begin{minipage}{.18\textwidth}
  	\centering
  	 ResNet18
    \includegraphics[ width=\linewidth]{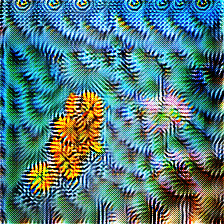}\
     Grey Owl
  \end{minipage}
  \begin{minipage}{.18\textwidth}
  	\centering
  	 ResNet50
    \includegraphics[ width=\linewidth]{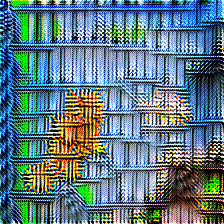}\
      Grey Owl
  \end{minipage}
    \begin{minipage}{.18\textwidth}
  	\centering
  	 ResNet101
    \includegraphics[ width=\linewidth]{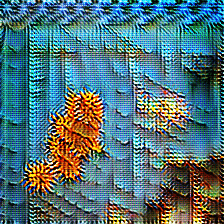}\
      Grey Owl
  \end{minipage}
\begin{minipage}{.18\textwidth}
  	\centering
  	 ResNet152
    \includegraphics[ width=\linewidth]{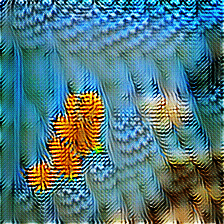}\
     Grey Owl
  \end{minipage}
  \caption{\emph{\textbf{Evolution of \textsc{TTP}:}} Unconstrained targeted adversarial patterns generated by \textsc{TTP} are shown to demonstrate how \textsc{TTP} evolves as it learns perturbations from different source models of a certain family of networks.}
\label{fig:sup_evolution_of_patterns_2}
\end{figure*}

Figures \ref{fig:1st_page_main_demo}, \ref{fig:2st_page_main_demo}, \ref{fig:3st_page_main_demo}, \ref{fig:4st_page_main_demo}, \ref{fig:5st_page_main_demo} and \ref{fig:6st_page_main_demo} show different targeted patterns produced by \textsc{TTP} trained against naturally trained ResNet50. We demonstrate how adversarial patterns evolve as \textsc{TTP} learns to model a certain target distribution from different networks of the same family in Figures \ref{fig:sup_evolution_of_patterns_1} and \ref{fig:sup_evolution_of_patterns_2}.

\end{document}